\pdfoutput=1

\documentclass[11pt]{article}

\usepackage[]{acl}

\usepackage{times}
\usepackage{latexsym}

\usepackage[T1]{fontenc}

\usepackage[utf8]{inputenc}

\usepackage{microtype}

\usepackage{times}
\usepackage{latexsym}
\usepackage{graphicx}
\usepackage{amsmath}
\usepackage{amssymb}
\usepackage{textcomp}
\usepackage{bbm}
\usepackage{bm}
\usepackage{multirow}
\usepackage{tabularx}
\usepackage{ragged2e}
\usepackage{booktabs}
\usepackage{subfig}
\usepackage{tablefootnote}
\usepackage{footnotehyper}
\usepackage{xspace}
\usepackage{enumitem}

\newcommand\mpm[0]{MPoSM\xspace}
\newcommand\mpmWoSpace[0]{MPoSM}

\title{\raisebox{-.3\height}{\includegraphics[height=0.8cm]{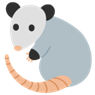}} Masked Part-Of-Speech Model: Does Modeling Long Context \\ Help Unsupervised POS-tagging?}

\author{Xiang Zhou\ \ \ \ \ \ Shiyue Zhang\ \ \ \ \ \ Mohit Bansal \\

  Department of Computer Science \\
University of North Carolina at Chapel Hill \\
  \texttt{\{xzh, shiyue, mbansal\}@cs.unc.edu} \\
}

\begin{document}
\maketitle

\begin{abstract}
Previous Part-Of-Speech (POS) induction models usually assume certain independence assumptions (e.g., Markov, unidirectional, local dependency) that do not hold in real languages. For example, the subject-verb agreement can be both long-term and bidirectional.
To facilitate flexible dependency modeling, we propose a Masked Part-of-Speech Model (\mpm), inspired by the recent success of Masked Language Models (MLM). \mpm can model arbitrary tag dependency and perform POS induction through the objective of masked POS reconstruction. We achieve competitive results on both the English Penn WSJ dataset as well as the universal treebank containing 10 diverse languages. 
Though modeling the long-term dependency should ideally help this task, our ablation study shows mixed trends in different languages.
To better understand this phenomenon, we design a novel synthetic experiment that can specifically diagnose the model's ability to learn tag agreement. Surprisingly, we find that even strong baselines fail to solve this problem consistently in a very simplified setting: the agreement between adjacent words. Nonetheless, \mpm achieves overall better performance. Lastly, we conduct a detailed error analysis to shed light on other remaining challenges.\footnote{Our code is available at \url{https://github.com/owenzx/MPoSM}} 

\end{abstract}

\section{Introduction}
\label{sec:intro}
\begin{figure}[t!]
    \centering
    \includegraphics[width=0.85\linewidth]{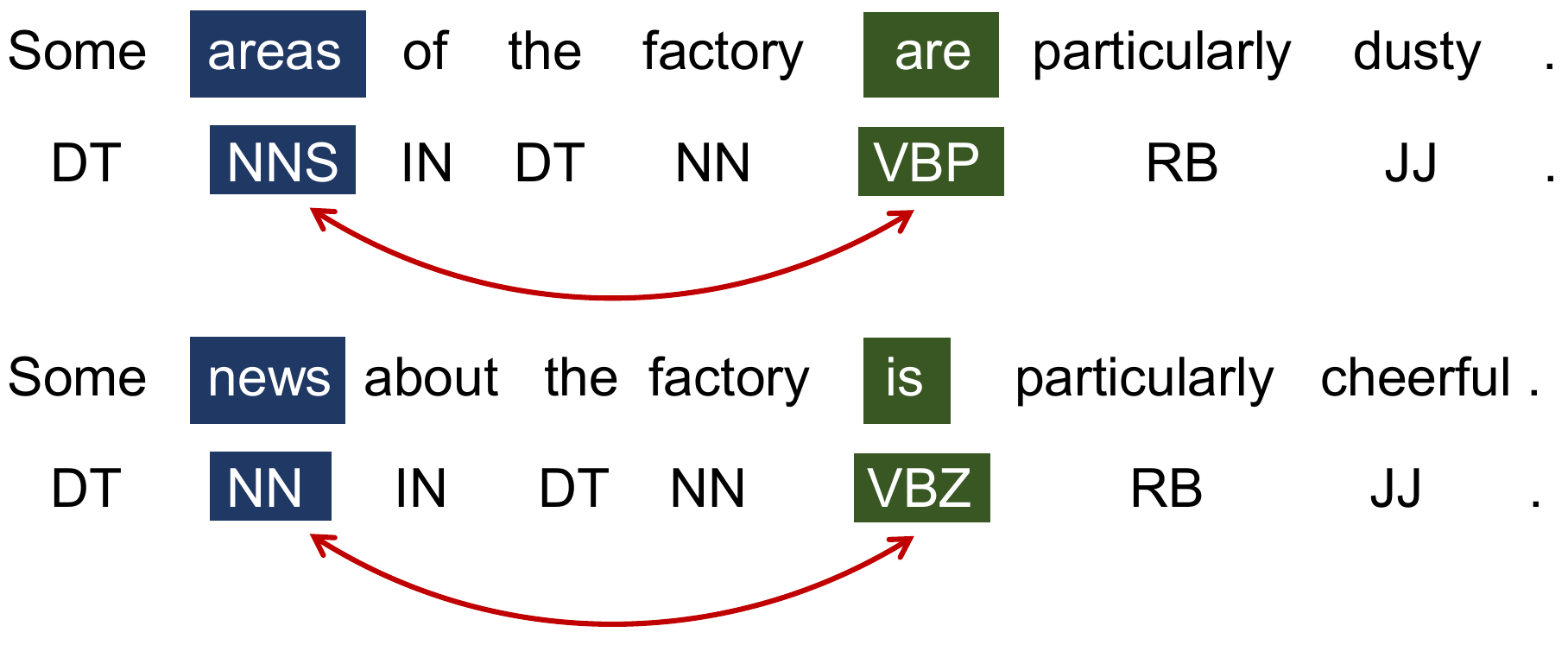}
    \vspace{-6pt}
    \caption{Two long-term tag dependency examples in English.}
    \vspace{-15pt}
    \label{fig:example}
\end{figure}

Unsupervised Part-Of-Speech (POS) tagging is the task of discovering POS tags from text without any supervision. 
These unsupervised syntax induction approaches can reduce the effort needed for collecting expensive syntactic annotation, and can bring us insights about what inductive bias leads to the emergence of syntax.
Recent POS induction models have made great progress using different frameworks \cite{christodoulopoulos2010two,berg2010painless,he2018unsupervised,stratos2016unsupervised,shi2020role}. However, most of them assume certain independence assumptions among POS tags, e.g., Markov~\cite{merialdo1994tagging,berg2010painless,ammar2014conditional,he2018unsupervised}, unidirectional~\cite{tran2016unsupervised}, local dependency~\cite{stratos2016unsupervised,gupta2020clustering}, etc. 
On the contrary, complex and long-term dependency appear in many real languages and plays an important role in defining the POS tags. 
For example, in Figure~\ref{fig:example}, the VBP tag of \textit{are} and the NNS tag of \textit{areas} depend on each other, and so do the VBZ tag of \textit{is} and the NN tag of \textit{news}.\footnote{Similar agreements are also common in many other languages. Various other long-term dependencies also exist, e.g., tense consistency, long-distance PP attachment, etc.} 
So in this case, models only conditioning the immediate preceding tag (Markov) or 1-2 neighboring words (local) cannot explain the distinction between NNS and NN, or between VBZ and VBP. While unidirectional (e.g., using a unidirectional LSTM~\cite{tran2016unsupervised}) models are in theory capable of modeling long-term dependency through optimizing the joint probability of tags, bidirectional architectures still show clear advantage in language modeling literature~\cite{bahdanau2015neural, vaswani2017attention, devlin2019bert}.

In this work, we present a novel framework for POS induction that is capable of modeling arbitrary long-term bidirectional dependencies: \textbf{M}asked \textbf{P}art-\textbf{O}f-\textbf{S}peech \textbf{M}odel (\textbf{\mpm}\footnote{\mpm is pronounced as m-possum (\raisebox{-.3\height}{\includegraphics[height=0.3cm]{possum.png}}).}), inspired by recent success of Masked Language Models (MLM) \cite{devlin2019bert}.
Specifically, \mpm consists of two modules (see Figure~\ref{fig:model_compare}): a \textit{local POS prediction} module that maps each word to its POS tag
and a \textit{masked POS reconstruction} module  
that masks a certain portion of tags produced from the previous step, and then learns to first predict the masked tags as latent variables and then reconstruct the corresponding words.
We use a bidirectional LSTM (Bi-LSTM) to predict the mask tags conditioned on the remaining tags,  which grants our model the ability to model complex long-term and bidirectional dependencies among tags. 
Through the training signal back-propagated from this module, the tags predicted from the \textit{local POS prediction} module will also be encouraged to have global inter-dependency, which leads to better tags.
Since we do not have gold POS tags, at the masked positions, we marginalize over all possible tags and optimize word reconstruction probabilities. 
Intuitively, the correct induction of POS tags is beneficial for the prediction of the correct masked words.
For example, in Figure~\ref{fig:example}, if we mask the second positions of the two sentences (corresponding to \textit{areas} and \textit{news}), 
inducing two different tags (i.e., NNS for \textit{areas} and NN for \textit{news}) correctly will make the word prediction easier than inducing the same tag.
From a probabilistic view, our model is conceptually similar to approximately modeling the probability of generating the whole sentence from latent tags using masked loss as a surrogate.

\mpm achieves competitive performance on both the 45-tag English Penn WSJ dataset~\cite{marcus-etal-1993-building} and the 12-tag universal treebank~\cite{mcdonald-etal-2013-universal} containing 10 diverse different languages. 
It achieves comparable oracle M-1 compared to the SOTA (state-of-the-art) models~\cite{stratos2019mutual, gupta2020clustering} on Penn WSJ dataset and achieves higher performance than~\citet{stratos2019mutual} on 4 out of 10 languages on the universal treebank.
We also show that substantial improvements can be made with the help of contextualized representations in mBERT, similar to \citet{gupta2020clustering}.
We conduct an ablation study on multiple languages by replacing the Bi-LSTM architecture with a window-based MLP that models the local dependency of tags.
Surprisingly, while modeling the full-sentence context can improve the performance of English and German, modeling local context is better for Indonesian and Korean. 
Our mutual information analysis indicates that this difference may be resulted from the different degrees of gold-tag dependency of different languages. 

Since real-life datasets can contain many confounding factors, we next design a suite of well-controlled synthetic experiments from the angle of agreement-learning to examine how well the model can take advantage of the long-term dependency. 
Our synthetic datasets can guarantee enough training signals for the model to capture the agreements. However, we show that all current models fail to consistently solve the agreement learning problems, even with the most basic agreement happening between adjacent words. Nonetheless, our model shows the best performance with the highest percentage of solving these problems in multiple runs. We conjecture that this is relate to the general optimization problems of latent variable inference problems \cite{jin2016local} (see more discussions in Section~\ref{sec:ana}). 
Such obstacles prevent models from gaining additional benefits from modeling long-term dependency. Finally, we did error analysis on the predicted clusters for English and Portuguese and identify remaining challenges both from imperfect modeling and lack of data diversity.

In summary, our main contributions are: (1) a novel POS induction architecture with MLM-inspired loss that allows learning arbitrary tag dependencies and reaches close-to-SOTA  performance; 
and (2) examining the effectiveness of using long-term context and providing a suite of synthetic datasets to expose the challenges in agreement learning and pointing out future challenges.

\section{Background}
\label{sec:bg}

\paragraph{POS Induction.} A POS tag is a category of words that share the same grammatical property. A simplified form of these tags will involve commonly known categories such as nouns, verbs, etc. Formally, given a sentence with $l$ words $\bm{x}=\{x_i\}_{i=1}^l$, the corresponding POS tags $\bm{z}=\{z_i\}_{i=1}^l$, then the goal of the POS induction task is to infer $\bm{z}$ from $\bm{x}$ without supervision from gold tags. 

\paragraph{Limitations of Existing POS Induction Models.} 
From the perspective of probabilistic graphical models, POS tags can be viewed as latent variables related to all the observed words. 
Each tag $z$ is a latent variable that generates the corresponding word $x$. Hence, inducing the POS tag sequence becomes the problem of performing MAP inference of the latent variables. 
This is a popular and effective view adopted by many previous works.
To make such inference tractable, previous works have to add certain assumptions, including adding Markov assumption to the latent variables $z$ (i.e., the current tag only depends on the \textit{immediate} previous tag)~\cite{merialdo1994tagging,berg2010painless,ammar2014conditional}, only considering local dependencies~\cite{stratos2019mutual,gupta2020clustering}, unidirectional dependencies~\cite{tran2016unsupervised}, etc.
However, dependencies in real language are not constrained by length or direction, as we discussed in Sec.~\ref{sec:intro}.
Hence, simplifying and ruling the capability out in the model design is suboptimal. To mitigate this problem, in Sec.~\ref{sec: model}, we will describe our approach to model long-term dependency.

\paragraph{Why are the Learned Latent Variables Correlated with POS-Tags?}  
Before introducing our method, we discuss why latent variable models can induce POS-tags well. 
Take the vanilla HMM as an example, the latent variables in the model can be viewed as being optimized towards two objectives: the transition probability $p(z_i|z_{i-1})$ and the emission probability $p(x_i|z_i)$. They characterize two properties respectively: (1) strong ordering dependencies among latent variables; and (2) the strong correlation between latent variables and the observed word. In short, the success of previous latent variable models implies: \textit{A word's inherent category that has strong ordering constraints will highly resemble the POS tag.}
In this work, we follow this assumption, but propose a model that is able to learn arbitrary bidirectional long-term dependencies $p(z_i|\{z_j\}_{j\neq i})$ instead of $p(z_i|z_{i-1})$.

\section{Masked Part-Of-Speech Model}
\label{sec: model}

Inspired by the recent success in masked language modeling (MLM)~\cite{devlin2019bert}, we present \textbf{M}asked \textbf{P}art-\textbf{O}f-\textbf{S}peech \textbf{M}odel (\textbf{\mpm}). 
Next, we will first describe the model architecture and then introduce several useful additional techniques.

\subsection{Model Architecture}
\label{sec:mpmarch}
\begin{figure}[t!]
    \centering
    \includegraphics[width=0.93\linewidth]{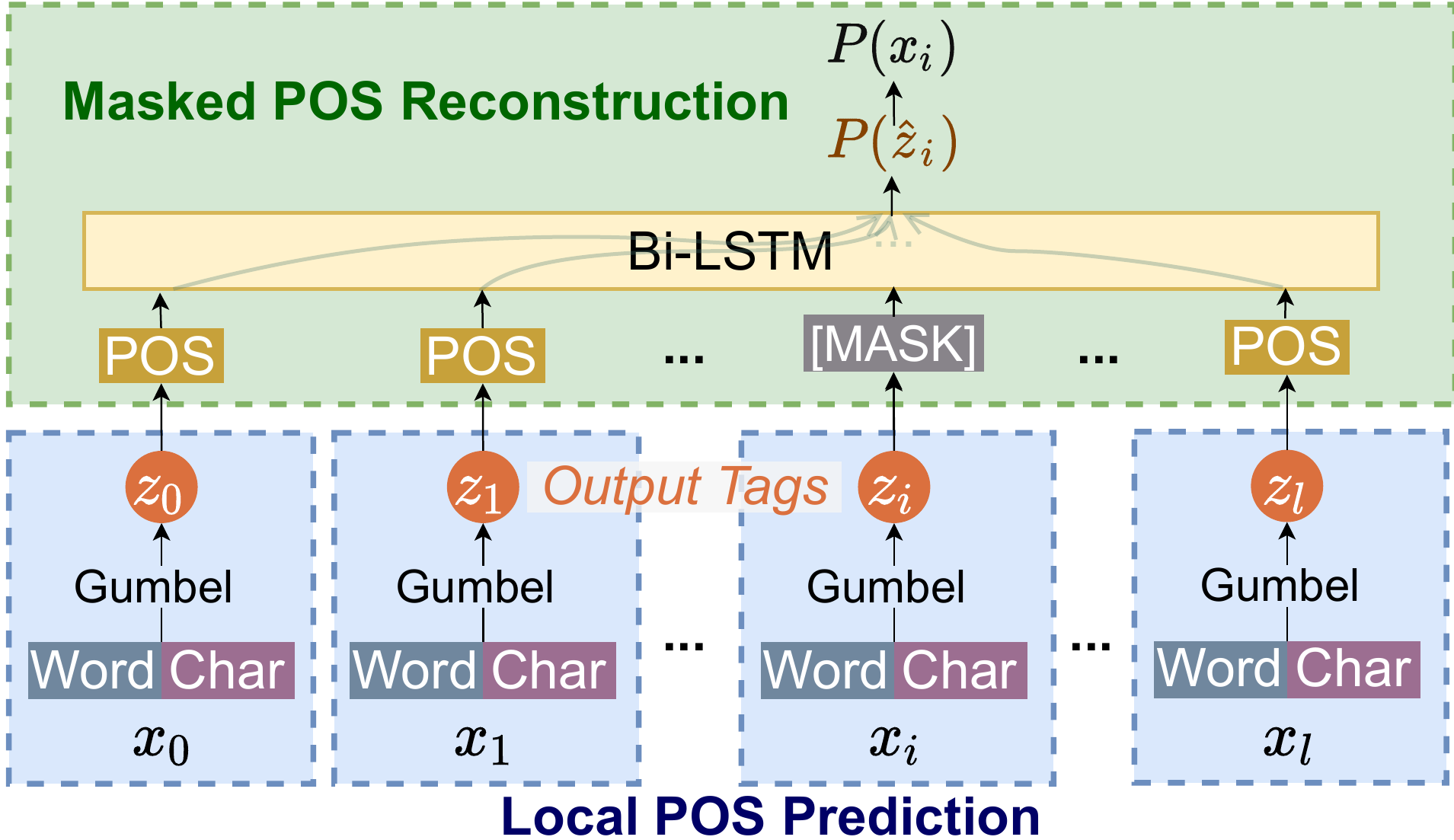}
    \vspace{-3pt}
    \caption{Illustration of our \mpm. The model consists of two parts: the \emph{local POS prediction} module (blue part at the bottom) and the \emph{masked POS reconstruction} module (green part at the top).}
    \vspace{-5pt}
    \label{fig:model_compare}
\end{figure}

As is shown in Figure~\ref{fig:model_compare}, our model consists of two parts: a \textit{local POS prediction} module and a \textit{masked POS reconstruction} module. 
The local POS prediction module predicts a POS tag for each word, and the masked POS reconstruction module encourages strong dependencies among these tags. 

\paragraph{Local POS Prediction.} 
Given the input word sequence $\bm{x}=\{x_i\}_{i=1}^l$ with length $l$, we first get the word embeddings.
As morphological features are shown to be useful for POS induction~\cite{christodoulopoulos2010two} to capture inflection (e.g., the `-s' suffix for English plurals), we follow \citet{stratos2019mutual} to extract character-level representations using a Bi-LSTM. We concatenate word embeddings and char representations to form the final representations for each word, $\bm{w}=\{w_i\}_{i=1}^l$.
Then, we use a single context-independent feed-forward network to predict the POS tags $\bm{z}$ out of $\bm{w}$, 
i.e. $z_i=\arg\max(\mathrm{Softmax}(\mathrm{FF}(w_i))$). 
Essentially, this module models $P(z_i|x_i)$ for every position and predicts the POS tag only conditioned on the word itself without considering its context.
We make this design choice as POS tags are the syntactic property of each \textit{individual word}, so it should be able to be predicted as an attribute of the word.\footnote{However, it gives our model the limitation of only predicting a fixed tag for each word, same as~\citet{stratos2019mutual}. Nonetheless, the upper bound M-1 on the 45-tag Penn dataset is 94.6 and the mean upper bound on UD is 95.4, which are both substantially higher than current models.} 
Importantly, in order to make the whole model end-to-end differentiable, we replace the $\arg\max$ with a straight-through Gumbel-Softmax estimator~\cite{JangGP17,MaddisonMT17} (see Appendix~\ref{appendix:gumbel} for more details). 

\paragraph{Masked POS Reconstruction.}
After we get all the predicted POS tags $\bm{z}=\{z_i\}_{i=1}^l$ for the previous module,
we conduct masked POS reconstruction to encourage modeling strong dependencies among $\bm{z}$. Specifically, we follow \citet{devlin2019bert} to mask $15\%$
of the predicted POS tags and replace them with a placeholder \texttt{MASK} tag. 
Then we map them into POS embeddings and use a Bi-LSTM~\cite{hochreiter1997long} as the dependency-modeling network.
\footnote{We also experimented using the Transformer architecture in our preliminary experiments, but did not observe additional gain. See Appendix~\ref{appendix:others} for details.} 
This grants flexibility of modeling the long-term and bidirectional dependency among tags without any assumptions and thus brings us an advantage over traditional HMM-based models. 
Then, we predict the masked POS tags out of the contextualized representations from the Bi-LSTM, so, essentially, it models $P(\hat{z}_j|C_j)$, where $\hat{z}_j$ is the reconstructed tag at position $j$ and $C_j=\{z_i\}_{i\neq j}$ is the context.
We treat the predicted $\hat{z}_j$ as latent variables and maximize the probability of the corresponding word $x_j$, which can be written out by marginalizing over $\hat{z}_j$: 
\vspace{-5pt}
\begin{equation}
P(x_j|C_j)=\sum_{\hat{z}_j} P(x_j|\hat{z}_j) P(\hat{z}_j|C_j)
\label{eqn:prob}
\vspace{-5pt}
\end{equation}
The conditional probability $P(x_j|\hat{z}_j)$ can be modeled through another feed-forward network with the POS embeddings as the input. 
Intuitively, predicting the $P(\hat{z}_j|C_j)$ objective encourages strong dependency among the tags and predicting $P(x_j|\hat{z}_j)$ reinforces the connection between the words and the tags.
Hence, the total loss is the sum of all the log-probabilities at masked positions:
\vspace{-5pt}
\begin{equation}
\mathcal{L}_{\text{MPoSM}} = \sum_{j\in \text{Masked\ Positions}}\log P(x_j|C_j) 
\label{eqn:loss}
\vspace{-5pt}
\end{equation}
Importantly, the supervision from this module will back-propagate to the \emph{local POS prediction} module. Therefore, even though it produces POS tags independently, the supervision helps it to capture the interdependency among all the tags.

During testing, we use the output of the local prediction module as the output tags.\footnote{We can also use the tags predicted in the masked POS reconstruction module $P(\hat{z}_i|C_i)$ as the output, but we find the output of local prediction module is empirically better.}

\subsection{Additional Techniques}
\label{sec:additional_tech}
Below, we introduce several additional techniques used in our model to achieve good performance.

\paragraph{Careful Initialization.}
Similar to many other unsupervised learning models~\cite{gimpel2012concavity,meila2013experimental,he2018unsupervised}, we found our model to be sensitive to initialization in our preliminary experiments. Below, we propose \textit{Masked Language Modeling Pretraining} (\textbf{MLMP}).
We use a two-stage training procedure: (1) we remove the modeling architecture for $P(x_j|\hat{z}_j)$ and $P(\hat{z}_j|C_j)$, and directly apply an MLP to model $P(x_j|C_j)$ without explicitly predicting the masked tag; (2) we initialize our \mpm with the pre-trained model in (1) and continue training with the loss in Eqn.~\ref{eqn:loss}. This procedure trains the bottom layers with a smoother objective and provides a better starting point for optimizing the \mpm loss.
Besides, the \mpm model can leverage knowledge from both \textit{pretrained embeddings} similar to \citet{he2018unsupervised} and  \citet{zhu-etal-2020-return}, or \textit{pretrained language models} similar to \citet{gupta2020clustering}.

\paragraph{Connecting P(x|z) and P(z|x).}
While the \emph{local POS prediction} module models $P(z|x)$, the \emph{masked POS reconstruction} module has a part that models $P(x|z)$ (Eqn.~\ref{eqn:prob}). These two probabilities can be connected using the Bayes' rule: $P(x|z) = \frac{P(z|x)P(x)}{\sum_{x}P(z|x)P(x)}$. If we assume the training set is representative enough of the language, we can approximate $P(x)$ by the word frequency in the dataset, and then we can compute $P(x|z)$ directly following the Bayes' rule instead of using a neural network to model it. We notice that binding these two probabilities can usually make the training more stable and improve the performance when training from scratch. 
Note that we do not adopt this change when using pretrained word embeddings because we can use the pretrained embedding weights at the output layer~\cite{press2017using}, which brings additional knowledge for the final word prediction.

\paragraph{Dataset Rechunking.}
One potential problem of using the full sentence context is the \textit{position bias of POS tags}. For example, since a large number of English sentences start with the word `the`, position 0 will have a strong bias towards predicting the `DT' tag. In our experiments, we concatenate all the sentences in the original dataset and re-chunk them randomly. Then we combine the rechunked dataset and the original dataset as our training set. In our preliminary experiments, we find it can improve the stability and the performance of the model.

\begin{table*}[t]
\centering
\small
\resizebox{0.9\textwidth}{!}{%
\begin{tabular}{l|c|cccccccccc}
\toprule
& en (Penn) & de & en (UD) & es & fr & id & it & ja & ko & pt-br & sv  \\
\midrule
\# sentences & \textbf{49208} & 15918 & \textbf{43948} & 16007 & 16422 & \underline{5593} & 7189 & 9494 & 6339 & 11998 & 6159 \\ 
\# words & \textbf{1173766} &  293460 & \textbf{1046829} & 424425 & 396511 & 121923 & 167873 & 92033 & \underline{69690} & 298323 & 96319 \\
\# vocab & 49206 & 52435 & 46348 & 50334 & 44453 & 22221 & 22344 & \textbf{56758} & 36335 & 34011 & \underline{16241} \\
Avg word freq. & \textbf{23.85} & 5.60 & \textbf{22.59} & 8.43 & 8.92 & 5.49 & 7.51 & \underline{1.62} & 1.92 & 8.77 & 5.93 \\
Tag Mutual Info. & \textbf{0.85} & 0.56 & \textbf{0.86} & 0.65 & 0.66 & 0.39 & 0.57 & 0.74 & \underline{0.27} & 0.59 & 0.59  \\

\bottomrule
\end{tabular}
}
\vspace{-5pt}
\caption{Dataset statistics. For each row, the language with the largest number are in \textbf{bold} and the language with the smallest number is \underline{underlined}. Computation details about the tag mutual information is in Appendix~\ref{appendix:mi}. 
}
\vspace{-5pt}
\label{tab:dataset}
\end{table*}

\section{Connections to Related Works}
The HMM-based POS induction model~\cite{merialdo1994tagging} has many extensions, including using hand-engineered linguistic features~\cite{berg2010painless}, pretrained embeddings~\cite{lin2015unsupervised}, task-specific modifications~\cite{blunsom2011hierarchical,stratos2016unsupervised}, flow-based transformations~\cite{he2018unsupervised}, external resources~\cite{haghighi2006prototype,snyder2008unsupervised,das2011unsupervised}, etc. They all optimize the probability of the sequence, $P(\bm{x})$. 
However, it requires certain dependency assumptions to be tractable.
Our model instead optimizes the sum of conditional word probabilities given the remaining context $\sum_i \log P(x_i|\bm{x}_{1..i-1}, \bm{x}_{i+1..l})$, i.e., MLM loss~\cite{devlin2019bert}. 
While being different from $P(\bm{x})$, this objective is an effective surrogate and makes modeling complex dependencies possible.
There also exist some earlier methods that do not require the Markov assumption. For example, \citet{abend-etal-2010-improved} design a method to directly cluster the embeddings containing distributional and morphological information of the word, and then identify prototype words to facilitate the final POS induction.  
\citet{tran2016unsupervised} propose a neural HMM model. Similar to our model, it can also model long-term dependency (due to the use of LSTM), however, they still constrain the dependence to be uni-directional (due to the HMM nature). Our model does not have such constraints and empirically works better.

Architecture-wise, our model is conceptually similar to a denoising auto-encoder~\cite{vincent2008extracting}, where the masking step can be viewed as adding noises to the POS tags. The idea of using auto-encoder models for unsupervised learning has been explored with CRFs in \citet{ammar2014conditional}. However, they still require Markov independence assumption to make inference on CRF tractable, while our model has the ability to model complex long-term dependencies. Plus, we use an MLM-inspired loss instead of reconstructing Brown clusters~\cite{brown1992class} as \citet{ammar2014conditional}. 

Our model also provide additional insight on the relation between MLM and syntax emergence. Such connection has also been explored in previous works.
Pretrained transformers using MLM~\cite{devlin2019bert,clark2019electra,raffel2020exploring} have shown strong syntactic abilities~\cite{tenney2019bert,jawahar2019does,goldberg2019assessing}. 
CBOW and skip-gram embeddings~\cite{tom2013effcient} can be viewed as an MLM with a limited context window (i.e., local context), and have been shown to be useful for syntax induction, especially with small window sizes~\cite{bansal2014tailoring, lin2015unsupervised, he2018unsupervised}. Some recent POS induction works explore CBOW-style objectives~\cite{stratos2019mutual,gupta2020clustering}.
However, using the sentence-level MLM objective for syntax induction is under-explored.
The only exception is the recent work of \citet{shen2020structformer}, which focuses on a different task: unsupervised parsing. The different tasks lead to substantially different focuses and designs in the architecture.
They use MLM with a dependency-constrained self-attention mechanism to extract parses, while we extend MLM to the POS-tag level (\mpm) and explicitly discretizes the latent variables to extract tags.

\section{Experimental Setup}
\subsection{Datasets and Metrics}
We evaluate our model on two datasets: the 45-tag English Penn WSJ dataset~\cite{marcus-etal-1993-building} and the 12-tag universal treebank~\cite{mcdonald-etal-2013-universal}. Following \citet{ammar2014conditional} and \citet{stratos2019mutual}, we use the v2.0 version\footnote{\url{https://github.com/ryanmcd/uni-dep-tb}} containing 10 different languages.
Detailed statistics are in Table~\ref{tab:dataset}.

Following recent works~\cite{stratos2019mutual,gupta2020clustering}, we use the Many-to-One accuracy (M-1)~\cite{johnson-2007-doesnt} as our metric, and train and evaluate our model on the whole dataset. 
Following~\citet{shi2020role}, we distinguish between the oracle performance that selects the model with the best M-1 metric (M-1\textsubscript{OR}), and the fully unsupervised performance that selects the model with the best loss (M-1). However, many previous works used different or unspecified model selection settings. For a fair comparison, we get results under our setting using their official code if possible.

\begin{table*}[t]
\centering
\small
\resizebox{0.88\textwidth}{!}{%
\begin{tabular}{l cccccccccc}
\toprule
& de & en & es & fr & id & it & ja & ko & pt-br & sv  \\
\midrule
\mpmWoSpace\textsubscript{OR} & 71.8 & 72.3 & 73.2 & 73.7 & 69.4 & 69.7 & 76.8 & 55.2 & 76.2 & 63.7  \\
\mpm + mBERT\textsubscript{OR} & 77.5 & 72.1 & 77.0 & 74.8 & 72.4 & 74.8 & 76.0 & 56.6 & 78.1 & 65.5  \\
\midrule
\citet{stratos2019mutual}\textsubscript{OR} & 75.4 & 73.1 & 73.1 & 70.4 & 73.6 & 67.4 & 77.9 & 65.6 & 70.7 & 67.1 \\
\citet{gupta2020clustering}\textsubscript{OR} & 81.7 & 76.7 & 79.5 & 70.8 & 76.9 & 71.8 & 84.7 & 69.7 & 78.9 & 69.7 \\
\midrule
\citet{stratos2016unsupervised}** & 63.4 & 71.4 & 74.3 & 71.9 & 67.3 & 60.2 & 69.4 & 61.8 & 65.8 & 61.0 \\
\citet{berg2010painless}** & 67.5 & 62.4 & 67.1 & 62.1 & 61.3 & 52.9 & 78.2 & 60.5 & 63.2 & 56.7 \\
\citet{brown1992class}** & 60.0 & 62.9 & 67.4 & 66.4 & 59.3 & 66.1 & 60.3 & 47.5 & 67.4 & 61.9 \\
\bottomrule
\end{tabular}
}
\vspace{-6pt}
\caption{Performance on the universal treebank. \citet{gupta2020clustering} also leverages pretrained mBERT model. All the other models do not use pretrained models or embeddings. Subscript \textsubscript{OR} denotes models evaluated by oracle M-1 and ** refers to unspecified model selection. Standard deviations and non-oracle numbers are in the Appendix~\ref{appendix:UD_res}. }
\vspace{-12pt}
\label{tab:ud}
\end{table*}

\subsection{Implementation Details}
For the English Penn WSJ dataset, we use the pretrained embedding provided in \citet{he2018unsupervised}. For the main results on the universal treebank, we do not use any external resources and use MLMP initialization. 
Additionally, we also report the results with mBERT contextualized representations on the universal treebank following \citet{gupta2020clustering}, where they show mBERT representations empirically outperforms English BERT representations on POS-tag induction.
Same to the implementation in \citet{gupta2020clustering}, we also use the average representation over all the subwords and layers as the representation for each word. 
For all models, we train our model using Adam~\cite{DBLP:journals/corr/KingmaB14} with an initial learning rate 0.001. The batch size is set to 80. 
The Gumbel softmax temperature is set to 2.0. 
The results on the Penn WSJ dataset are the mean of 5 runs, and the results on the universal treebank are the mean of 3 runs (see more details in Appendix~\ref{sec:app_detail}).

\begin{table}[t]
\centering
\small
\resizebox{0.48\textwidth}{!}{
\begin{tabular}{lcc}
\toprule
&  M-1\textsubscript{OR} & M-1  \\
\midrule
\mpm & 75.6 ($\pm$2.0) & 74.5 ($\pm$1.4) \\
\mpm + emb & 78.6  ($\pm$1.7)  & 77.9  ($\pm$1.8) \\
\midrule
\citet{tran2016unsupervised} & - & 75.0  ($\pm$ 1.5) \\
\citet{he2018unsupervised} & - & 75.6 ($\pm$2.7)\\
\citet{stratos2019mutual} & 78.1  ($\pm$0.8)  & -\\
\citet{gupta2020clustering} & - &  79.5 ($\pm$0.9) \\
\midrule
\citet{brown1992class}** & \multicolumn{2}{c}{65.6} \\
\citet{berg2010painless}** & \multicolumn{2}{c}{74.9 ($\pm$1.5)} \\
\citet{tran2016unsupervised}** & \multicolumn{2}{c}{79.1}  \\
\citet{abend-etal-2010-improved}** &  \multicolumn{2}{c}{75.1}\\
\citet{he2018unsupervised}* &  - & 80.8 ($\pm$1.3)\\
\bottomrule
\end{tabular}
}
\vspace{-5pt}
\caption{POS induction performance on the 45-tag English Penn WSJ dataset. 
Numbers are the 5-run averages plus standard deviations.
In the last row group, we include the numbers of baselines that have unspecified model selection procedures and no official code available (denoted by **), or use a more carefully designed model selection method (denoted by *).
}
\vspace{-5pt}
\label{tab:main}
\end{table}

\section{Results and Ablations}
\label{sec:results}

\subsection{45-tag English Penn WSJ dataset.} 
The results are shown in Table~\ref{tab:main}. We reported two variants: the \mpm model that does not use any external resource, and the \mpm + emb model that uses the same pretrained word embeddings as \citet{he2018unsupervised}. Using pretrained embeddings does provide substantial improvements to our model.
Overall, our model achieves competitive performance compared to SOTA models~\cite{stratos2019mutual,gupta2020clustering}, reaching 78.6 oracle M-1. 
The oracle performance is 0.5 points higher than the model in \citet{stratos2019mutual} using a mutual information-based loss.
Our fully unsupervised performance reaches 77.9 M-1, which is also similar to SOTA models~\cite{stratos2019mutual,gupta2020clustering}, 
and is higher compared to previous models using the same pretrained embeddings~\cite{he2018unsupervised} (75.6), models not using the Markov assumption~\cite{abend-etal-2010-improved} (75.1) or models using uni-directional long-term dependency~\cite{tran2016unsupervised} (75.0). 
Concurrent to our work, \citet{gupta2020clustering} achieve a higher M-1 of 79.5, but they use more additional resources, including mBERT representations and \texttt{fastText}~\cite{joulin-etal-2017-bag} morphological features.

\subsection{12-tag Multilingual Results on Universal Treebank.}
We also report results on all 10 languages on the universal treebank in Table~\ref{tab:ud} (the full table with standard deviations can be found in Table~\ref{tab:ud_app} of Appendix~\ref{appendix:UD_res}). To make the settings practical to low-resource languages, we do not use any pretrained word embeddings on this dataset. 
Compared to the SOTA model \cite{stratos2019mutual} that also does not use any external resources, our model achieves competitive performance, outperforming it on 4 out of 10 languages (es, fr, it, pt-br). 
Together with the English Penn WSJ dataset, we notice that \mpm perform well on most of the languages, but may underperform the previous model on languages with weaker tag-level dependency (e.g., ko and id, statstics are in Table~\ref{tab:dataset}, more detailed analyses and discussions are in Appendix~\ref{appendix:mi} and \ref{appendix:korean}) and on smaller datasets (e.g., ko and sv).

Concurrently, \citet{gupta2020clustering} showed substantial improvement on the universal treebank by leveraging knowledge in the pretrained mBERT representations. Inspired by their success, we also report the result using mBERT in \mpm (as denoted by \mpm + mBERT) in Table~\ref{tab:ud}. Similarly, using mBERT substantially improves \mpm's performance on the universal treebank. While on languages with weak tag-level dependency or smaller datasets, \mpm + mBERT still does not perform most effectively (similar to the trend in \mpm), \mpm + mBERT achieves substantially higher results on most of the languages compared to \mpm, and achieves results higher than \citet{gupta2020clustering} on French and Italian. On average, \citet{gupta2020clustering} still achieves higher results. 
This trend may imply that other factors (e.g. the clustering methods used in \citet{gupta2020clustering}) are important for their good performance. 
We have also tried using mBERT on the English WSJ dataset, but do not see a substantial improvement. 
We leave how to combine their method with \mpm as a promising future direction.

\begin{table}[t]
\centering
\small
\resizebox{0.48\textwidth}{!}{
\begin{tabular}{l cccc}
\toprule
& en (Penn) & de (uni) & id  (uni) & ko  (uni)\\
\midrule
\mpm (full) & \textbf{78.6} ($\pm$1.7)& \textbf{71.8}  ($\pm$2.5)&  69.4 ($\pm$1.8)& 55.2 ($\pm$1.3)\\
\mpm (width=2) & 77.3 ($\pm$0.3) & 68.5 ($\pm$2.8) &  \textbf{70.0} ($\pm$1.0) & \textbf{56.6} ($\pm$1.4) \\
\bottomrule
\end{tabular}
}
\vspace{-5pt}
\caption{Oracle M-1 performance of different context types on the four different languages.}
\vspace{-5pt}
\label{tab:context}
\end{table}

\subsection{How does modeling long context influence the results?} 
Taking advantage of the flexibility of our model, we analyze whether modeling long-term context is always better than modeling local context.
We compare two models: the \mpm (full) model is the default model described in Sec.~\ref{sec: model}, and the \mpm (width=2) model that replaces the Bi-LSTM network with an MLP and only takes in the neighboring 4 predicted POS tags as the input (i.e., local context).
We test our model on 4 languages: English, German, Indonesian, and Korean. These languages are selected to have representative statistics among all the languages in the universal treebank in terms of dataset size and average word frequency (see Table~\ref{tab:dataset}).\footnote{We choose language mainly according to the dataset statistics instead of linguistic properties as in preliminary experiments, those statistics are more influential to the performance.}
The results are in Table~\ref{tab:context}. 
On English and German, the default model is better than the \mpm (width=2) variant by 1.3 and 3.3 points respectively. However, on Indonesian and Korean, the trend is reversed with the \mpm (width=2) variant showing 0.6 and 1.4 point advantage respectively. 
We notice that the languages do not benefit from using a longer context also correlates well with the languages with weak tag-level dependencies. Such property prevent the \mpm from benefit from the advantage of dependence modeling on those languages, and consequently using a longer context does not provide additional help. More detailed analysis is in Appendix~\ref{appendix:mi}.

\section{Analysis and Challenges}
\label{sec:ana}

\subsection{Agreement Learning Experiments}
\label{sec:agreement}
Inducing good POS tags requires models to understand what ``agreement'' is. To match the gold 45-tag set of Penn Treebank, the model needs to distinguish between VBP (Verb, non-3rd person singular present) and VBZ (Verb, 3rd person singular present) tags (see examples in Figure~\ref{fig:example}). Though local morphological features do provide useful cues for such classification, models should achieve better performance by observing the full picture of agreement in the long context.
From the results in Sec.~\ref{sec:results}, we notice evaluation in real-life datasets may contain many confounding factors.
Hence, we design a well-controlled synthetic dataset to examine exactly how well the model learns these agreements.
Surprisingly, we find that the limitation of current models is not about leveraging long context, but a common fundamental limitation on using co-dependency to distinguish tags.

\begin{figure}[t!]
    \centering
    \includegraphics[width=0.88\linewidth]{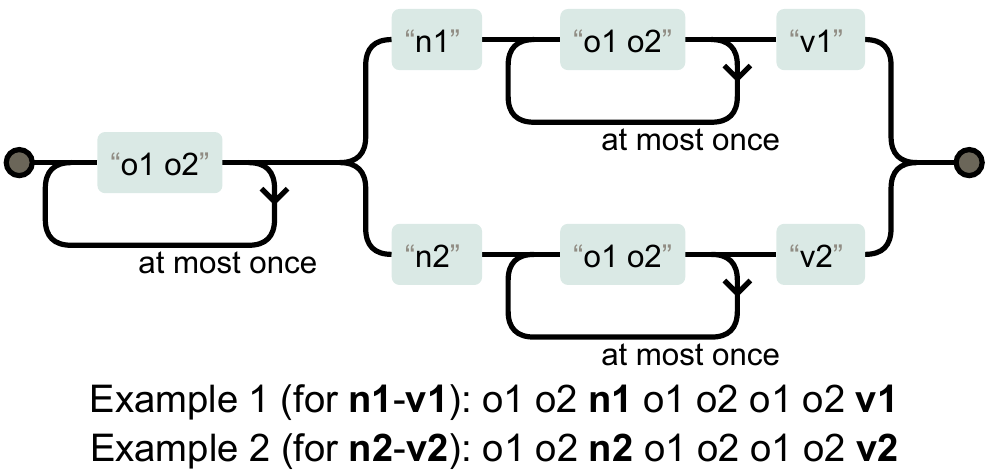}
    \vspace{-3pt}
    \caption{Illustration of the tag-level regex for sentences in  D(2-4). D(0) sentences can be generated by removing the ``\texttt{o1} \texttt{o2}'' block between \texttt{n} and \texttt{v}.}
    \label{fig:syn_re}
    \vspace{-6pt}
\end{figure}

\paragraph{Controlled Dataset Design.}
To provide a simplified and well-controlled environment, our synthetic datasets consist of a small set of words and tags, and simple sentences. Specifically, we use 6 different tags, with 5 unique words correspond to each tag. 
Our 6 tags are named after nouns (\texttt{n1}, \texttt{n2}), verbs (\texttt{v1}, \texttt{v2}), and other unimportant tokens (\texttt{o1}, \texttt{o2}).
In every sentence, \texttt{n1} always appear before \texttt{v1}, and \texttt{n2} before \texttt{v2}, analogous to subject-verb agreement in English. 
We create the synthetic data by first sampling a tag sequence (illustrated in Figure~\ref{fig:syn_re}) and then randomly select words of each tag in the sequence. We also make sure the two agreements (\texttt{n1}-\texttt{v1}, \texttt{n2}-\texttt{v2}) have exactly symmetric data.
We use the ``distance'', i.e., the number of tokens, between \texttt{n} and \texttt{v} to control the agreement-learning difficulty. The larger the distance is, the harder the example is.
Therefore, we create two subsets with different levels of difficulty, and each contains 40,000 sentences.
In the first simpler subset D(0), \texttt{n} and \texttt{v} are adjacent. In the second harder subset D(2-4), \texttt{n} and \texttt{v} are separated by 2-4 words. 
Complete illustrations, regexes and  additional results are in the Appendix~\ref{sec:app_agreement}.

\begin{table}[t]
\centering
\small
\resizebox{0.44\textwidth}{!}{
\begin{tabular}{l ccc}
\toprule
& D(0) & D(2-4)   \\
\midrule
\mpm (width=1) & \textbf{99.50} (95\%) & 87.19 (0\%) \\
\mpm (width=2) & 92.99 (30\%) & 87.62 (0\%) \\
\mpm (full) & 96.50 (75\%) & \textbf{95.31} (30\%) \\
\midrule
\citet{stratos2019mutual} (width=2) & 92.99 (30\%) & 86.56 (0\%) \\
\citet{tran2016unsupervised} (full) & 80.97 (0\%) & 82.50 (0\%) \\
\bottomrule
\end{tabular}
}
\vspace{-5pt}
\caption{Oracle M-1 on the synthetic datasets with the percentage of perfect runs (100 M-1) in the bracket.}
\vspace{-9pt}
\label{tab:syn_result}
\end{table}

\paragraph{Surprising Difficulty of Learning Agreement.}
Model performances on our synthetic datasets are in Table~\ref{tab:syn_result}. We report the mean M-1 of 20 runs and the percentage of perfect runs (achieving 100 M-1), as models are expected to consistently achieve the perfect score if they really acquire agreement. We include three variants of \mpm using different contexts (from the \textit{width=1} model only considering the immediate neighboring tokens to the \textit{full} model considering the whole sentence). 
We compare with two representative baselines: the SOTA model \citet{stratos2019mutual} which uses the context with width=2, and the neural HMM model~\cite{tran2016unsupervised} which leverages unidirectional full context. 
In Table~\ref{tab:syn_result}, we first notice the surprising difficulty of learning agreement even in the simple D(0) setting, where the \texttt{n} and \texttt{v} are already adjacent. None of the models can consistently produce the perfect tags in this setting. The best results are from \mpm with a specific inductive bias of only using the width=1 local context, but it still fails to achieve the perfect score consistently.  
Other models using a larger context show substantially lower results.
On the harder D(2-4) setting, we see similar observations. Due to the architecture limitation, none of the models using local context can achieve the perfect score even once. Models using the long context also fail to perform well consistently. \mpm (full) is the single best model that successfully acquires agreement, albeit only 30\% of the time.
These observations demonstrate the difficulty of learning agreement in POS induction. As reflected by the results on D(0), such difficulty cannot be fully attributed to the long-term issue. We suspect the latent variable-based loss function used in current models can contain many bad local minimums, similar to the optimization difficulty observed in Gaussian Mixture Models~\cite{jin2016local}. Models are likely to stuck in one local minimum (e.g., viewing \texttt{n1} and \texttt{n2} as the same tag) and never reach the global optimum.

Finally, we want to point out that our findings are not contradictory with recent studies that show the derivation of agreements from MLM-style models~\cite{jawahar2019does,goldberg2019assessing,lin-etal-2019-open}. One key difference is that they directly measure the word-level agreement, e.g., \emph{are} should follow \emph{areas} in Figure~\ref{fig:example}. However, POS induction focuses on the tag-level agreement, i.e., VBP should follow NNS. 
Our \mpm can also be viewed as adding an explicit discretization step in a normal MLM so that we can predict discrete tags.
If we remove the POS-factorization step in Eqn~\ref{eqn:prob}, and directly predict the word from the word context, our model can also capture the word-level agreement.

\subsection{Error Analysis of Predicted Clusters}
In Table~\ref{tab:main}, we notice that performances of different models are saturating around 78 M-1 on the English Penn WSJ dataset. 
To examine the limitations of current models point out future directions, we manually investigated the clusters learned by our model. 
Below, we list our main findings on English (see similar findings of Portuguese in Appendix~\ref{appendix:portuguese}):

\begin{figure}[t!]
    \centering
    \includegraphics[width=0.98\linewidth]{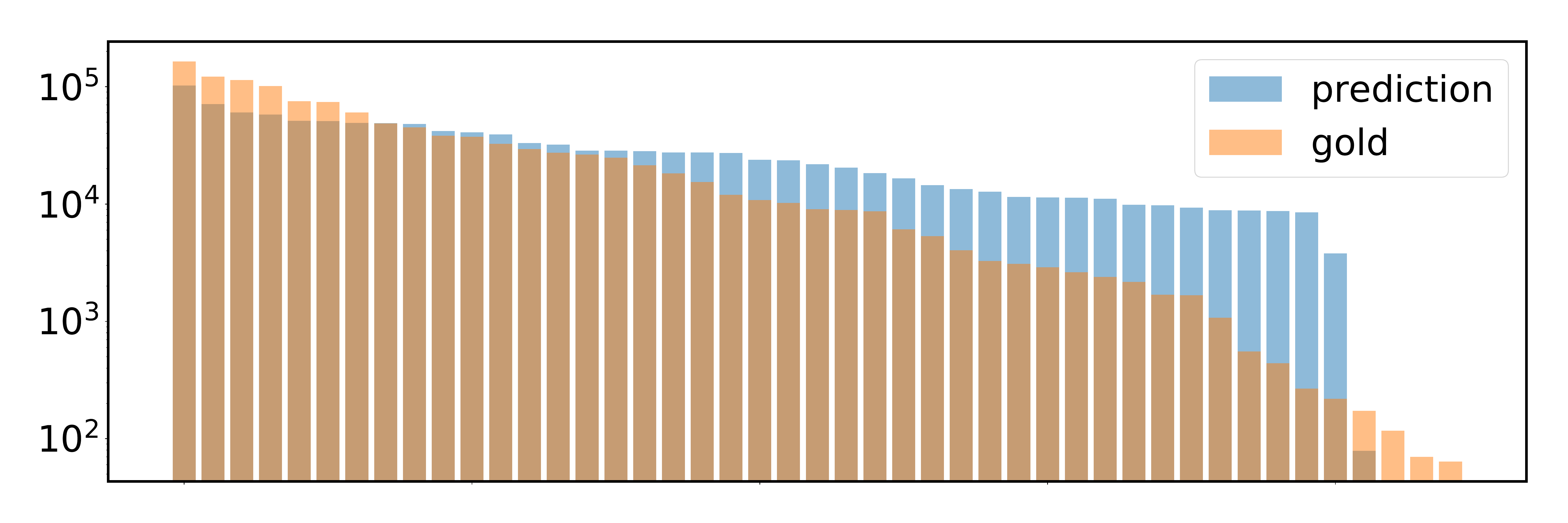}
    \caption{Log-scale sizes of the predicted clusters and the gold clusters.}
    \vspace{-6pt}
    \label{fig:size}
    \vspace{-6pt}
\end{figure}

\paragraph{The sizes of predicted clusters are more uniform than gold clusters.}
Only 1 predicted cluster contains very few (less than 3000) words, while the scale of gold clusters showing a much larger variance, with 29\% of the 45 clusters containing less than 3000 words. 
A bar plot illustration is in Figure~\ref{fig:size}. We can see that the size of gold clusters has a much larger range than the predicted clusters.
Under the current losses, assigning a small number of words to one tag is likely to make the loss worse, but it hard to match the skewed distribution of natural tags. \citet{johnson-2007-doesnt} show similar findings on HMM models trained with EM. These consistent findings may hint at a common limitation of current objectives. Future work should explore different objectives with more suitable inductive biases. 

\paragraph{Agreements are not learned well.}
Similar to the observation in Sec.~\ref{sec:agreement}, agreements are not learned well in the predicted clusters. For example, the VBP tag (Verb, non-3rd person singular present) is an important tag in the subject-verb agreement. While this tag has 15377 occurrences in the gold annotations, it is not the major tag in any of the predicted clusters. Most VBP words are either mixed with the VB or the VBZ words. We consistently observe models fail to separate these verbs, showing a large room for improvement.

\paragraph{Difficulty in mapping one word to multiple tags.}
Without using mBERT representations, \mpm (also applies to many other models, e.g., \citet{he2018unsupervised,stratos2019mutual}, etc.) predicts the same tag for one word. However, the same word can have different tags in different contexts. 
For example, the word `that' can have gold tags IN, RB, and WDT.  
Future works should explore directions on capturing the multi-sense phenomenon.

\paragraph{Dataset biases influence predicted clusters.}
For example, the English WSJ dataset contains many financial news reports, so numerical words (e.g., `million', `billion') and  related symbols like `\%'  are very frequent. Since these words always appear in a distinctive context, models will naturally cluster these tokens together. Hence, we encourage future research to explore more diverse datasets.

\section{Conclusion}
We propose \mpm, a POS induction model inspired by MLM and can model complex long-term dependencies between POS tags. 
Our model shows competitive performance on both English and multilingual datasets. We analyze the effectiveness of using long context compared to local context. Finally, we use synthetic datasets and analyses to point out remaining challenges.

\section{Ethical Considerations}
The model proposed in this paper is intended to study how syntax emerge from unsupervised learning objectives. It can also help understand languages with limited annotations. However, as we showed in this paper, the syntax predicted by current models can contain errors and be influenced by the choice of datasets, so the model's output should be used with caution and examined by experts. 
Our model has been tested on 10 diverse languages. Our findings on these languages should generalize to languages with similar linguistic properties, but we suggest careful empirical studies before applying our approach to languages distant from those we study in this paper.

\section*{Acknowledgments}
We thank the reviewers for their helpful comments. We thank Chao Lou for the help in data preprocessing steps. This work was supported by ONR Grant N00014-18-1-2871, NSF-CAREER Award 1846185, DARPA MCS Grant N66001-19-2-4031, and a Bloomberg Data Science Ph.D. Fellowship.
The views are those of the authors and not of the funding agency.

\bibliography{custom}

\begin{thebibliography}{46}
\expandafter\ifx\csname natexlab\endcsname\relax\def\natexlab#1{#1}\fi

\bibitem[{Abend et~al.(2010)Abend, Reichart, and
  Rappoport}]{abend-etal-2010-improved}
Omri Abend, Roi Reichart, and Ari Rappoport. 2010.
\newblock \href {https://aclanthology.org/P10-1132} {Improved unsupervised
  {POS} induction through prototype discovery}.
\newblock In \emph{Proceedings of the 48th Annual Meeting of the Association
  for Computational Linguistics}, pages 1298--1307, Uppsala, Sweden.
  Association for Computational Linguistics.

\bibitem[{Ammar et~al.(2014)Ammar, Dyer, and Smith}]{ammar2014conditional}
Waleed Ammar, Chris Dyer, and Noah~A. Smith. 2014.
\newblock \href
  {https://proceedings.neurips.cc/paper/2014/hash/b9f94c77652c9a76fc8a442748cd54bd-Abstract.html}
  {Conditional random field autoencoders for unsupervised structured
  prediction}.
\newblock In \emph{Advances in Neural Information Processing Systems 27: Annual
  Conference on Neural Information Processing Systems 2014, December 8-13 2014,
  Montreal, Quebec, Canada}, pages 3311--3319.

\bibitem[{Bahdanau et~al.(2015)Bahdanau, Cho, and Bengio}]{bahdanau2015neural}
Dzmitry Bahdanau, Kyunghyun Cho, and Yoshua Bengio. 2015.
\newblock \href {http://arxiv.org/abs/1409.0473} {Neural machine translation by
  jointly learning to align and translate}.
\newblock In \emph{3rd International Conference on Learning Representations,
  {ICLR} 2015, San Diego, CA, USA, May 7-9, 2015, Conference Track
  Proceedings}.

\bibitem[{Bansal et~al.(2014)Bansal, Gimpel, and Livescu}]{bansal2014tailoring}
Mohit Bansal, Kevin Gimpel, and Karen Livescu. 2014.
\newblock \href {https://doi.org/10.3115/v1/P14-2131} {Tailoring continuous
  word representations for dependency parsing}.
\newblock In \emph{Proceedings of the 52nd Annual Meeting of the Association
  for Computational Linguistics (Volume 2: Short Papers)}, pages 809--815,
  Baltimore, Maryland. Association for Computational Linguistics.

\bibitem[{Berg-Kirkpatrick et~al.(2010)Berg-Kirkpatrick, Bouchard-C{\^o}t{\'e},
  DeNero, and Klein}]{berg2010painless}
Taylor Berg-Kirkpatrick, Alexandre Bouchard-C{\^o}t{\'e}, John DeNero, and Dan
  Klein. 2010.
\newblock \href {https://aclanthology.org/N10-1083} {Painless unsupervised
  learning with features}.
\newblock In \emph{Human Language Technologies: The 2010 Annual Conference of
  the North {A}merican Chapter of the Association for Computational
  Linguistics}, pages 582--590, Los Angeles, California. Association for
  Computational Linguistics.

\bibitem[{Blunsom and Cohn(2011)}]{blunsom2011hierarchical}
Phil Blunsom and Trevor Cohn. 2011.
\newblock \href {https://aclanthology.org/P11-1087} {A hierarchical
  {P}itman-{Y}or process {HMM} for unsupervised part of speech induction}.
\newblock In \emph{Proceedings of the 49th Annual Meeting of the Association
  for Computational Linguistics: Human Language Technologies}, pages 865--874,
  Portland, Oregon, USA. Association for Computational Linguistics.

\bibitem[{Brown et~al.(1992)Brown, Della~Pietra, Desouza, Lai, and
  Mercer}]{brown1992class}
Peter~F Brown, Vincent~J Della~Pietra, Peter~V Desouza, Jennifer~C Lai, and
  Robert~L Mercer. 1992.
\newblock Class-based n-gram models of natural language.
\newblock \emph{Computational linguistics}, 18(4):467--480.

\bibitem[{Christodoulopoulos et~al.(2010)Christodoulopoulos, Goldwater, and
  Steedman}]{christodoulopoulos2010two}
Christos Christodoulopoulos, Sharon Goldwater, and Mark Steedman. 2010.
\newblock \href {https://aclanthology.org/D10-1056} {Two decades of
  unsupervised {POS} induction: How far have we come?}
\newblock In \emph{Proceedings of the 2010 Conference on Empirical Methods in
  Natural Language Processing}, pages 575--584, Cambridge, MA. Association for
  Computational Linguistics.

\bibitem[{Clark et~al.(2020)Clark, Luong, Le, and Manning}]{clark2019electra}
Kevin Clark, Minh{-}Thang Luong, Quoc~V. Le, and Christopher~D. Manning. 2020.
\newblock \href {https://openreview.net/forum?id=r1xMH1BtvB} {{ELECTRA:}
  pre-training text encoders as discriminators rather than generators}.
\newblock In \emph{8th International Conference on Learning Representations,
  {ICLR} 2020, Addis Ababa, Ethiopia, April 26-30, 2020}. OpenReview.net.

\bibitem[{Das and Petrov(2011)}]{das2011unsupervised}
Dipanjan Das and Slav Petrov. 2011.
\newblock \href {https://aclanthology.org/P11-1061} {Unsupervised
  part-of-speech tagging with bilingual graph-based projections}.
\newblock In \emph{Proceedings of the 49th Annual Meeting of the Association
  for Computational Linguistics: Human Language Technologies}, pages 600--609,
  Portland, Oregon, USA. Association for Computational Linguistics.

\bibitem[{Devlin et~al.(2019)Devlin, Chang, Lee, and
  Toutanova}]{devlin2019bert}
Jacob Devlin, Ming-Wei Chang, Kenton Lee, and Kristina Toutanova. 2019.
\newblock \href {https://doi.org/10.18653/v1/N19-1423} {{BERT}: Pre-training of
  deep bidirectional transformers for language understanding}.
\newblock In \emph{Proceedings of the 2019 Conference of the North {A}merican
  Chapter of the Association for Computational Linguistics: Human Language
  Technologies, Volume 1 (Long and Short Papers)}, pages 4171--4186,
  Minneapolis, Minnesota. Association for Computational Linguistics.

\bibitem[{Gimpel and Smith(2012)}]{gimpel2012concavity}
Kevin Gimpel and Noah~A. Smith. 2012.
\newblock \href {https://aclanthology.org/N12-1069} {Concavity and
  initialization for unsupervised dependency parsing}.
\newblock In \emph{Proceedings of the 2012 Conference of the North {A}merican
  Chapter of the Association for Computational Linguistics: Human Language
  Technologies}, pages 577--581, Montr{\'e}al, Canada. Association for
  Computational Linguistics.

\bibitem[{Goldberg(2019)}]{goldberg2019assessing}
Yoav Goldberg. 2019.
\newblock \href {https://arxiv.org/abs/1901.05287} {Assessing bert's syntactic
  abilities}.
\newblock \emph{ArXiv preprint}, abs/1901.05287.

\bibitem[{Gupta et~al.(2022)Gupta, Shi, Gimpel, and
  Sachan}]{gupta2020clustering}
Vikram Gupta, Haoyue Shi, Kevin Gimpel, and Mrinmaya Sachan. 2022.
\newblock Deep clustering of text representations for supervision-free probing
  of syntax.
\newblock In \emph{Proceedings of the AAAI Conference on Artificial
  Intelligence}.

\bibitem[{Haghighi and Klein(2006)}]{haghighi2006prototype}
Aria Haghighi and Dan Klein. 2006.
\newblock \href {https://aclanthology.org/N06-1041} {Prototype-driven learning
  for sequence models}.
\newblock In \emph{Proceedings of the Human Language Technology Conference of
  the {NAACL}, Main Conference}, pages 320--327, New York City, USA.
  Association for Computational Linguistics.

\bibitem[{He et~al.(2018)He, Neubig, and Berg-Kirkpatrick}]{he2018unsupervised}
Junxian He, Graham Neubig, and Taylor Berg-Kirkpatrick. 2018.
\newblock \href {https://doi.org/10.18653/v1/D18-1160} {Unsupervised learning
  of syntactic structure with invertible neural projections}.
\newblock In \emph{Proceedings of the 2018 Conference on Empirical Methods in
  Natural Language Processing}, pages 1292--1302, Brussels, Belgium.
  Association for Computational Linguistics.

\bibitem[{Hochreiter and Schmidhuber(1997)}]{hochreiter1997long}
Sepp Hochreiter and J{\"u}rgen Schmidhuber. 1997.
\newblock Long short-term memory.
\newblock \emph{Neural computation}, 9(8):1735--1780.

\bibitem[{Jang et~al.(2017)Jang, Gu, and Poole}]{JangGP17}
Eric Jang, Shixiang Gu, and Ben Poole. 2017.
\newblock \href {https://openreview.net/forum?id=rkE3y85ee} {Categorical
  reparameterization with gumbel-softmax}.
\newblock In \emph{5th International Conference on Learning Representations,
  {ICLR} 2017, Toulon, France, April 24-26, 2017, Conference Track
  Proceedings}. OpenReview.net.

\bibitem[{Jawahar et~al.(2019)Jawahar, Sagot, and Seddah}]{jawahar2019does}
Ganesh Jawahar, Beno{\^\i}t Sagot, and Djam{\'e} Seddah. 2019.
\newblock \href {https://doi.org/10.18653/v1/P19-1356} {What does {BERT} learn
  about the structure of language?}
\newblock In \emph{Proceedings of the 57th Annual Meeting of the Association
  for Computational Linguistics}, pages 3651--3657, Florence, Italy.
  Association for Computational Linguistics.

\bibitem[{Jiang et~al.(2016)Jiang, Han, and Tu}]{jiang-etal-2016-unsupervised}
Yong Jiang, Wenjuan Han, and Kewei Tu. 2016.
\newblock \href {https://doi.org/10.18653/v1/D16-1073} {Unsupervised neural
  dependency parsing}.
\newblock In \emph{Proceedings of the 2016 Conference on Empirical Methods in
  Natural Language Processing}, pages 763--771, Austin, Texas. Association for
  Computational Linguistics.

\bibitem[{Jin et~al.(2016)Jin, Zhang, Balakrishnan, Wainwright, and
  Jordan}]{jin2016local}
Chi Jin, Yuchen Zhang, Sivaraman Balakrishnan, Martin~J. Wainwright, and
  Michael~I. Jordan. 2016.
\newblock \href
  {https://proceedings.neurips.cc/paper/2016/hash/3875115bacc48cca24ac51ee4b0e7975-Abstract.html}
  {Local maxima in the likelihood of gaussian mixture models: Structural
  results and algorithmic consequences}.
\newblock In \emph{Advances in Neural Information Processing Systems 29: Annual
  Conference on Neural Information Processing Systems 2016, December 5-10,
  2016, Barcelona, Spain}, pages 4116--4124.

\bibitem[{Johnson(2007)}]{johnson-2007-doesnt}
Mark Johnson. 2007.
\newblock \href {https://aclanthology.org/D07-1031} {Why doesn{'}t {EM} find
  good {HMM} {POS}-taggers?}
\newblock In \emph{Proceedings of the 2007 Joint Conference on Empirical
  Methods in Natural Language Processing and Computational Natural Language
  Learning ({EMNLP}-{C}o{NLL})}, pages 296--305, Prague, Czech Republic.
  Association for Computational Linguistics.

\bibitem[{Joulin et~al.(2017)Joulin, Grave, Bojanowski, and
  Mikolov}]{joulin-etal-2017-bag}
Armand Joulin, Edouard Grave, Piotr Bojanowski, and Tomas Mikolov. 2017.
\newblock \href {https://aclanthology.org/E17-2068} {Bag of tricks for
  efficient text classification}.
\newblock In \emph{Proceedings of the 15th Conference of the {E}uropean Chapter
  of the Association for Computational Linguistics: Volume 2, Short Papers},
  pages 427--431, Valencia, Spain. Association for Computational Linguistics.

\bibitem[{Kingma and Ba(2015)}]{DBLP:journals/corr/KingmaB14}
Diederik~P. Kingma and Jimmy Ba. 2015.
\newblock \href {http://arxiv.org/abs/1412.6980} {Adam: {A} method for
  stochastic optimization}.
\newblock In \emph{3rd International Conference on Learning Representations,
  {ICLR} 2015, San Diego, CA, USA, May 7-9, 2015, Conference Track
  Proceedings}.

\bibitem[{Lin et~al.(2015)Lin, Ammar, Dyer, and Levin}]{lin2015unsupervised}
Chu-Cheng Lin, Waleed Ammar, Chris Dyer, and Lori Levin. 2015.
\newblock \href {https://doi.org/10.3115/v1/N15-1144} {Unsupervised {POS}
  induction with word embeddings}.
\newblock In \emph{Proceedings of the 2015 Conference of the North {A}merican
  Chapter of the Association for Computational Linguistics: Human Language
  Technologies}, pages 1311--1316, Denver, Colorado. Association for
  Computational Linguistics.

\bibitem[{Lin et~al.(2019)Lin, Tan, and Frank}]{lin-etal-2019-open}
Yongjie Lin, Yi~Chern Tan, and Robert Frank. 2019.
\newblock \href {https://doi.org/10.18653/v1/W19-4825} {Open sesame: Getting
  inside {BERT}{'}s linguistic knowledge}.
\newblock In \emph{Proceedings of the 2019 ACL Workshop BlackboxNLP: Analyzing
  and Interpreting Neural Networks for NLP}, pages 241--253, Florence, Italy.
  Association for Computational Linguistics.

\bibitem[{Maddison et~al.(2017)Maddison, Mnih, and Teh}]{MaddisonMT17}
Chris~J. Maddison, Andriy Mnih, and Yee~Whye Teh. 2017.
\newblock \href {https://openreview.net/forum?id=S1jE5L5gl} {The concrete
  distribution: {A} continuous relaxation of discrete random variables}.
\newblock In \emph{5th International Conference on Learning Representations,
  {ICLR} 2017, Toulon, France, April 24-26, 2017, Conference Track
  Proceedings}. OpenReview.net.

\bibitem[{Marcus et~al.(1993)Marcus, Santorini, and
  Marcinkiewicz}]{marcus-etal-1993-building}
Mitchell~P. Marcus, Beatrice Santorini, and Mary~Ann Marcinkiewicz. 1993.
\newblock \href {https://aclanthology.org/J93-2004} {Building a large annotated
  corpus of {E}nglish: The {P}enn {T}reebank}.
\newblock \emph{Computational Linguistics}, 19(2):313--330.

\bibitem[{McDonald et~al.(2013)McDonald, Nivre, Quirmbach-Brundage, Goldberg,
  Das, Ganchev, Hall, Petrov, Zhang, T{\"a}ckstr{\"o}m, Bedini,
  Bertomeu~Castell{\'o}, and Lee}]{mcdonald-etal-2013-universal}
Ryan McDonald, Joakim Nivre, Yvonne Quirmbach-Brundage, Yoav Goldberg, Dipanjan
  Das, Kuzman Ganchev, Keith Hall, Slav Petrov, Hao Zhang, Oscar
  T{\"a}ckstr{\"o}m, Claudia Bedini, N{\'u}ria Bertomeu~Castell{\'o}, and
  Jungmee Lee. 2013.
\newblock \href {https://aclanthology.org/P13-2017} {{U}niversal {D}ependency
  annotation for multilingual parsing}.
\newblock In \emph{Proceedings of the 51st Annual Meeting of the Association
  for Computational Linguistics (Volume 2: Short Papers)}, pages 92--97, Sofia,
  Bulgaria. Association for Computational Linguistics.

\bibitem[{Meila and Heckerman(1998)}]{meila2013experimental}
Marina Meila and David Heckerman. 1998.
\newblock \href
  {https://dslpitt.org/uai/displayArticleDetails.jsp?mmnu=1\&smnu=2\&article\_id=271\&proceeding\_id=14}
  {An experimental comparison of several clustering and initialization
  methods}.
\newblock \emph{{UAI} '98: Proceedings of the Fourteenth Conference on
  Uncertainty in Artificial Intelligence}, pages 386--395.

\bibitem[{Merialdo(1994)}]{merialdo1994tagging}
Bernard Merialdo. 1994.
\newblock \href {https://aclanthology.org/J94-2001} {Tagging {E}nglish text
  with a probabilistic model}.
\newblock \emph{Computational Linguistics}, 20(2):155--171.

\bibitem[{Mikolov et~al.(2013)Mikolov, Chen, Corrado, and
  Dean}]{tom2013effcient}
Tom{\'{a}}s Mikolov, Kai Chen, Greg Corrado, and Jeffrey Dean. 2013.
\newblock \href {http://arxiv.org/abs/1301.3781} {Efficient estimation of word
  representations in vector space}.
\newblock In \emph{1st International Conference on Learning Representations,
  {ICLR} 2013, Scottsdale, Arizona, USA, May 2-4, 2013, Workshop Track
  Proceedings}.

\bibitem[{Press and Wolf(2017)}]{press2017using}
Ofir Press and Lior Wolf. 2017.
\newblock \href {https://aclanthology.org/E17-2025} {Using the output embedding
  to improve language models}.
\newblock In \emph{Proceedings of the 15th Conference of the {E}uropean Chapter
  of the Association for Computational Linguistics: Volume 2, Short Papers},
  pages 157--163, Valencia, Spain. Association for Computational Linguistics.

\bibitem[{Raffel et~al.(2020)Raffel, Shazeer, Roberts, Lee, Narang, Matena,
  Zhou, Li, and Liu}]{raffel2020exploring}
Colin Raffel, Noam Shazeer, Adam Roberts, Katherine Lee, Sharan Narang, Michael
  Matena, Yanqi Zhou, Wei Li, and Peter~J Liu. 2020.
\newblock Exploring the limits of transfer learning with a unified text-to-text
  transformer.
\newblock \emph{Journal of Machine Learning Research}, 21(140):1--67.

\bibitem[{Shen et~al.(2021)Shen, Tay, Zheng, Bahri, Metzler, and
  Courville}]{shen2020structformer}
Yikang Shen, Yi~Tay, Che Zheng, Dara Bahri, Donald Metzler, and Aaron
  Courville. 2021.
\newblock \href {https://doi.org/10.18653/v1/2021.acl-long.559}
  {{S}truct{F}ormer: Joint unsupervised induction of dependency and
  constituency structure from masked language modeling}.
\newblock In \emph{Proceedings of the 59th Annual Meeting of the Association
  for Computational Linguistics and the 11th International Joint Conference on
  Natural Language Processing (Volume 1: Long Papers)}, pages 7196--7209,
  Online. Association for Computational Linguistics.

\bibitem[{Shi et~al.(2020)Shi, Livescu, and Gimpel}]{shi2020role}
Haoyue Shi, Karen Livescu, and Kevin Gimpel. 2020.
\newblock \href {https://doi.org/10.18653/v1/2020.emnlp-main.614} {On the role
  of supervision in unsupervised constituency parsing}.
\newblock In \emph{Proceedings of the 2020 Conference on Empirical Methods in
  Natural Language Processing (EMNLP)}, pages 7611--7621, Online. Association
  for Computational Linguistics.

\bibitem[{Snyder et~al.(2008)Snyder, Naseem, Eisenstein, and
  Barzilay}]{snyder2008unsupervised}
Benjamin Snyder, Tahira Naseem, Jacob Eisenstein, and Regina Barzilay. 2008.
\newblock \href {https://aclanthology.org/D08-1109} {Unsupervised multilingual
  learning for {POS} tagging}.
\newblock In \emph{Proceedings of the 2008 Conference on Empirical Methods in
  Natural Language Processing}, pages 1041--1050, Honolulu, Hawaii. Association
  for Computational Linguistics.

\bibitem[{Srivastava et~al.(2014)Srivastava, Hinton, Krizhevsky, Sutskever, and
  Salakhutdinov}]{srivastava2014dropout}
Nitish Srivastava, Geoffrey Hinton, Alex Krizhevsky, Ilya Sutskever, and Ruslan
  Salakhutdinov. 2014.
\newblock Dropout: a simple way to prevent neural networks from overfitting.
\newblock \emph{The journal of machine learning research}, 15(1):1929--1958.

\bibitem[{Stratos(2019)}]{stratos2019mutual}
Karl Stratos. 2019.
\newblock \href {https://doi.org/10.18653/v1/N19-1113} {Mutual information
  maximization for simple and accurate part-of-speech induction}.
\newblock In \emph{Proceedings of the 2019 Conference of the North {A}merican
  Chapter of the Association for Computational Linguistics: Human Language
  Technologies, Volume 1 (Long and Short Papers)}, pages 1095--1104,
  Minneapolis, Minnesota. Association for Computational Linguistics.

\bibitem[{Stratos et~al.(2016)Stratos, Collins, and
  Hsu}]{stratos2016unsupervised}
Karl Stratos, Michael Collins, and Daniel Hsu. 2016.
\newblock \href {https://doi.org/10.1162/tacl_a_00096} {Unsupervised
  part-of-speech tagging with anchor hidden {M}arkov models}.
\newblock \emph{Transactions of the Association for Computational Linguistics},
  4:245--257.

\bibitem[{Tenney et~al.(2019)Tenney, Das, and Pavlick}]{tenney2019bert}
Ian Tenney, Dipanjan Das, and Ellie Pavlick. 2019.
\newblock \href {https://doi.org/10.18653/v1/P19-1452} {{BERT} rediscovers the
  classical {NLP} pipeline}.
\newblock In \emph{Proceedings of the 57th Annual Meeting of the Association
  for Computational Linguistics}, pages 4593--4601, Florence, Italy.
  Association for Computational Linguistics.

\bibitem[{Tran et~al.(2016)Tran, Bisk, Vaswani, Marcu, and
  Knight}]{tran2016unsupervised}
Ke~M. Tran, Yonatan Bisk, Ashish Vaswani, Daniel Marcu, and Kevin Knight. 2016.
\newblock \href {https://doi.org/10.18653/v1/W16-5907} {Unsupervised neural
  hidden {M}arkov models}.
\newblock In \emph{Proceedings of the Workshop on Structured Prediction for
  {NLP}}, pages 63--71, Austin, TX. Association for Computational Linguistics.

\bibitem[{Vaswani et~al.(2017)Vaswani, Shazeer, Parmar, Uszkoreit, Jones,
  Gomez, Kaiser, and Polosukhin}]{vaswani2017attention}
Ashish Vaswani, Noam Shazeer, Niki Parmar, Jakob Uszkoreit, Llion Jones,
  Aidan~N. Gomez, Lukasz Kaiser, and Illia Polosukhin. 2017.
\newblock \href
  {https://proceedings.neurips.cc/paper/2017/hash/3f5ee243547dee91fbd053c1c4a845aa-Abstract.html}
  {Attention is all you need}.
\newblock In \emph{Advances in Neural Information Processing Systems 30: Annual
  Conference on Neural Information Processing Systems 2017, December 4-9, 2017,
  Long Beach, CA, {USA}}, pages 5998--6008.

\bibitem[{Vincent et~al.(2008)Vincent, Larochelle, Bengio, and
  Manzagol}]{vincent2008extracting}
Pascal Vincent, Hugo Larochelle, Yoshua Bengio, and Pierre{-}Antoine Manzagol.
  2008.
\newblock \href {https://doi.org/10.1145/1390156.1390294} {Extracting and
  composing robust features with denoising autoencoders}.
\newblock In \emph{Machine Learning, Proceedings of the Twenty-Fifth
  International Conference {(ICML} 2008), Helsinki, Finland, June 5-9, 2008},
  volume 307 of \emph{{ACM} International Conference Proceeding Series}, pages
  1096--1103. {ACM}.

\bibitem[{Yang et~al.(2019)Yang, Liu, Xie, Wang, and
  Balasubramanian}]{yang-etal-2019-latent}
Xuewen Yang, Yingru Liu, Dongliang Xie, Xin Wang, and Niranjan Balasubramanian.
  2019.
\newblock \href {https://doi.org/10.18653/v1/D19-1072} {Latent part-of-speech
  sequences for neural machine translation}.
\newblock In \emph{Proceedings of the 2019 Conference on Empirical Methods in
  Natural Language Processing and the 9th International Joint Conference on
  Natural Language Processing (EMNLP-IJCNLP)}, pages 780--790, Hong Kong,
  China. Association for Computational Linguistics.

\bibitem[{Zhu et~al.(2020)Zhu, Bisk, and Neubig}]{zhu-etal-2020-return}
Hao Zhu, Yonatan Bisk, and Graham Neubig. 2020.
\newblock \href {https://doi.org/10.1162/tacl_a_00337} {The return of lexical
  dependencies: Neural lexicalized {PCFG}s}.
\newblock \emph{Transactions of the Association for Computational Linguistics},
  8:647--661.

\end{thebibliography}
\bibliographystyle{acl_natbib}

\appendix

\section{Straight-Through Gubmbel-Softmax Estimator}
\label{appendix:gumbel}
To allow the gradients from the \emph{masked POS construction} module to back-propagate to the \emph{local POS prediction} module, we replace the standard argmax in the local POS prediction module with the \emph{straight-through Gumbel-Softmax} estimator. Specifically, we follow \citet{JangGP17,MaddisonMT17} to calculate the one-hot POS-tag vector using $\mathtt{one\_hot}(softmax(g_j+\mathit{logit}_j, \tau))$. In this equation, $\tau$ is the softmax temperature and $g_j$ are i.i.d. samples drawn from a standard Gumbel distribution, i.e., $g_j=-\log(-\log(U(0,1)))$, where $U(0,1)$ is a uniform distribution over the range $[0,1]$. Following \citet{JangGP17}, we use $\arg\max$ to discretize the distribution to a one-hot vector in the forward pass, but back-propagate through the continuous Gumbel-softmax. With this technique, the whole \mpm becomes end-to-end differentiable.

Alternatively, instead of operating on the single POS embedding of the Gumbel-Softmax output, we can also use the weighted sum of all the POS embeddings with weight $P(z_i|x_i)$. However, empirically, we notice the weighted sum approach does not perform well when the number of tags is large (e.g., 45 in the Penn WSJ dataset).
\begin{table*}[t]
\centering
\small
\resizebox{0.98\textwidth}{!}{
\begin{tabular}{l cccccccccc}
\toprule
& de & en & es & fr & id & it & ja & ko & pt-br & sv  \\
\midrule
\mpmWoSpace\textsubscript{OR} & 71.8 & 72.3 & 73.2 & 73.7 & 69.4 & 69.7 & 76.8 & 55.2 & 76.2 & 63.7  \\
& ($\pm$2.5) & ($\pm$1.7) & ($\pm$1.7) & ($\pm$1.3) & ($\pm$1.8) & ($\pm$3.6) & ($\pm$1.5) & ($\pm$1.3) & ($\pm$0.2) & ($\pm$2.6) \\
\mpm & 68.3 & 70.0 & 69.7 & 71.7 & 67.8 & 64.2 & 74.7 & 52.7 & 74.7 & 61.9  \\
& ($\pm$2.9) & ($\pm$2.2) & ($\pm$4.5) & ($\pm$2.3) & ($\pm$0.8) & ($\pm$1.9) & ($\pm$0.7) & ($\pm$0.4) & ($\pm$0.7) & ($\pm$1.9) \\
\mpm + mBERT\textsubscript{OR} & 77.5 & 72.1 & 77.0 & 74.8 & 72.4 & 74.8 & 76.0 & 56.6 & 78.1 & 65.5  \\
& ($\pm$0.3) & ($\pm$1.5) & ($\pm$1.5) & ($\pm$4.2) & ($\pm$5.2) & ($\pm$1.0) & ($\pm$1.1) & ($\pm$1.3) & ($\pm$1.7) & ($\pm$4.6) \\
\mpm + mBERT & 75.8 & 68.5 & 75.4 & 73.6 & 71.0 & 73.4 & 73.3 & 55.1 & 77.4 & 64.2  \\
& ($\pm$0.8) & ($\pm$3.6) & ($\pm$1.3) & ($\pm$5.3) & ($\pm$5.2) & ($\pm$1.0) & ($\pm$1.4) & ($\pm$1.6) & ($\pm$2.3) & ($\pm$3.5) \\
\midrule
\citet{stratos2019mutual}\textsubscript{OR} & 75.4 & 73.1 & 73.1 & 70.4 & 73.6 & 67.4 & 77.9 & 65.6 & 70.7 & 67.1 \\
& ($\pm$1.5) & ($\pm$1.7) & ($\pm$1.0) & ($\pm$2.9) & ($\pm$1.5) & ($\pm$3.3) & ($\pm$0.4) & ($\pm$1.2) & ($\pm$2.3) & ($\pm$1.5) \\
\citet{gupta2020clustering}\textsubscript{OR} & 81.7 & 76.7 & 79.5 & 70.8 & 76.9 & 71.8 & 84.7 & 69.7 & 78.9 & 69.7 \\
\midrule
\citet{stratos2016unsupervised}** & 63.4 & 71.4 & 74.3 & 71.9 & 67.3 & 60.2 & 69.4 & 61.8 & 65.8 & 61.0 \\
\citet{berg2010painless}** & 67.5 & 62.4 & 67.1 & 62.1 & 61.3 & 52.9 & 78.2 & 60.5 & 63.2 & 56.7 \\
& ($\pm$1.8) & ($\pm$3.5) & ($\pm$3.1) & ($\pm$4.5) & ($\pm$3.9) & ($\pm$2.9) & ($\pm$2.9) & ($\pm$3.6) & ($\pm$2.2) & ($\pm$2.5) \\
\citet{brown1992class}** & 60.0 & 62.9 & 67.4 & 66.4 & 59.3 & 66.1 & 60.3 & 47.5 & 67.4 & 61.9 \\
& ($\pm$1.8) & ($\pm$1.0) & ($\pm$1.7) & ($\pm$4.0) & ($\pm$1.2) & ($\pm$3.9) & ($\pm$1.2) & ($\pm$1.5) & ($\pm$1.8) & ($\pm$2.8) \\
\bottomrule
\end{tabular}
}
\vspace{-3pt}
\caption{Performance on the Universal Dependency dataset. \citet{gupta2020clustering} also leverage pretrained mBERT model. All the other models do not use pretrained models or embeddings. Subscript \textsubscript{OR} denotes models evaluated by oracle M-1 and ** refers to unspecified model selection.}
\vspace{-3pt}
\label{tab:ud_app}
\end{table*}

\section{Dataset Links}
The universal treebank dataset is from \url{https://github.com/ryanmcd/uni-dep-tb}. The English Penn WSJ dataset can be obtained through LDC.

\begin{table*}[t]
\centering
\small
\begin{tabular}{l cccccccccc}
\toprule
Context Type $C_i$ & de & en & es & fr & id & it & ja & ko & pt-br & sv  \\
\midrule
$z_{i-2}, z_{i-1}$ & 0.56 & 0.86 & 0.65 & 0.66 & 0.39 & 0.57 & 0.74 & 0.27 & 0.59 & 0.59  \\
$z_{i+2}, z_{i+1}$ & 0.56 & 0.86 & 0.65 & 0.67 & 0.39 & 0.58 & 0.70 & 0.23 & 0.59 & 0.58  \\
$z_{i+2}, z_{i+1}, z_{i-1}, z_{i-2}$ & 1.30 & 1.92 & 1.36 & 1.38 & 1.02 & 1.32 & 2.09 & 0.86 & 1.28 & 1.51  \\

\bottomrule
\end{tabular}
\vspace{-3pt}
\caption{Mutual information between the tag-level context and the tag $z_i$ on all 10 languages in the universal treebank.}
\vspace{-10pt}
\label{tab:ud_mi}
\end{table*}

\section{Implementation and Hyper-Parameter Details}
\label{sec:app_detail}
For initialization, for the English 45-tag Penn WSJ dataset, we use the pretrained word embedding provided in \citet{he2018unsupervised}. For the main results on the universal treebank, we do not use any external resources and always initialize our models using MLM pretraining. 
Additionally, we also report the results with mBERT contextualized representations on the universal treebank following \citet{gupta2020clustering}. 
Same to the implementation in \citet{gupta2020clustering}, we also use the average representation over all the subwords and layers as the representation for each word. 
We apply the ``connecting P(x|z) and P(z|x)'' technique for all our models not using pretrained word embeddings or pretrained language models. We apply the dataset rechunking technique to all our experiments.

For the hyper-parameters, we train all our model using Adam~\cite{DBLP:journals/corr/KingmaB14} with an initial learning rate 0.001. The batch size is set to 80 and we decay the learning rate with a factor 0.1 the loss stagnates. We set the word embedding dimension to 100, POS embedding dimension to 200, the character embedding dimension to 100, and the hidden vector dimension to 128. We use one layer of MLP in the \emph{local POS prediction} module and one layer of Bi-LSTM in the \emph{masked POS reconstruction} module. The masking rate is set to 15\% and the Gumbel softmax temperature is set to 2.0. We set the dropout rate~\cite{srivastava2014dropout} to 0.5. Specially, with pretraining word embeddings, we tie the input and output embeddings following~\citep{press2017using} and add one more layer in the local POS prediction layer to more effectively convert the pretrained embedding to POS tags following~\cite{he2018unsupervised}. And for our synthetic experiments, since the vocabulary size is small, we use a smaller character embedding dimension of 8. We use the loss as the metric to judge if our model has been converged. 
In this work, the results on the Penn WSJ dataset are the mean of 5 different runs, and the results on the universal treebank are the mean of 3 different runs. The experiments using mBERT can be run on a single RTX A6000 GPU, and all other experiments can be conducted on a single TITAN~Xp GPU. The time of experiments can take from several hours to several days, depending on the size of the dataset and the models.

\section{Results and Standard Deviations on the Universal Treebank}
\label{appendix:UD_res}
The full means and standard deviations (for our model and for previous works that reported this number) are shown in Table~\ref{tab:ud_app}. We also include the fully unsupervised performance (evaluating the model with the best loss) in the \mpm row. Our fully unsupervised model is slightly worse than the oracle version of both our model and \citet{stratos2019mutual}, but show comparable or higher performance to other results~\cite{stratos2016unsupervised,berg2010painless,brown1992class}.

\begin{table*}[t]
\centering
\small
\begin{tabular}{l cl}
\toprule
Name & Distance between \texttt{n} and \texttt{v} & Regex  \\
\midrule
D(0) & 0 & (o1 o2)\{1,2\}(n1 v1 | n2 v2)\\
MORPH & 0 & same as D(0) (+ morph. feature) \\
D(2-4) & 2-4 & (o1 o2)\{1,2\}(n1 (o1 o2)\{1,2\} v1 | n2 (o1 o2)\{1,2\} v2) \\
\bottomrule
\end{tabular}
\vspace{-3pt}
\caption{Tag-level regular expressions and the distances between \texttt{n} and \texttt{v} for each synthetic subset.}
\vspace{-3pt}
\label{tab:syn_design_app}
\end{table*}

\section{Analysis on the Dependency among Gold Tags}
\label{appendix:mi}
In Sec.~\ref{sec:results}, we notice that \mpm does not work equally well on all the languages. For example, in Table~\ref{tab:context}, we can see that out of 4 different languages, using full context instead of local context only improve 2 of them: English and German. In this section, we provide evidence that these different trends can result from the different strength of dependencies among tags in different languages. 

Assuming a tag sequence is $z_1, z_2, \ldots, z_n$, we compute the mutual information between a tag-level context of $z_i$ (denoted as $C_i$) and the tag $z_i$. A larger mutual information value can represent stronger dependencies among the gold tags.\footnote{Note that using a larger context will always lead to a larger mutual information value due to the property of mutual information. However, directly comparing the mutual information value with a very long context is confounded by many spurious correlations in the dataset. Hence, in this study, we only compare mutual information value in a limited context. Nonetheless, the trend shown in Table~.\ref{tab:ud_mi} is consistent across different context types.} The results are shown in Table~\ref{tab:ud_mi}. We can see that for all kinds of mutual information calculated in the table, Korean and Indonesian has the two lowest values, both substantially lower than the value of German and English. Notably, Korean and Indonesian are also the worst two languages of the \mpm model, while German and English and two of the languages with better performances. 
By its design, our model will induce tags that have strong dependencies among each other (see Sec.~\ref{sec:bg}). Hence, it is not strange that on Korean and Indonesian, the \mpm model could produce tags different from the gold tags and become less effective. And due to the difference between the predicted tags and the gold tags, it is not surprising to see that using a larger context in these two languages does not help the \mpm model in these two languages.

\begin{table}[t]
\centering
\small
\begin{tabular}{l cccc}
\toprule
& en (Penn) &  ko  (uni)\\
\midrule
\mpm (full) & \textbf{78.6} ($\pm$1.7)& 55.2 ($\pm$1.3)\\
\mpm (width=2) & 77.3 ($\pm$0.3) & 56.6 ($\pm$1.4) \\
\mpm-Word (full) & 72.2 ($\pm$1.6)& 54.1 ($\pm$3.3)\\
\mpm-Word (width=2) & 69.5 ($\pm$0.4)& \textbf{62.2} ($\pm$0.7)\\
\bottomrule
\end{tabular}
\vspace{-3pt}
\caption{Oracle M-1 Performance of different context types on English and Korean.}
\vspace{-3pt}
\label{tab:context_app}
\end{table}

\section{Discussion on uni-Korean Results}
\label{appendix:korean}
In Sec.~\ref{sec:results}, we notice that \mpm does not work well on the Korean dataset in the universal treebank. Inspired by \citet{stratos2019mutual} that achieves decent performance on Korean and the observation in Appendix~\ref{appendix:mi}, we study another modification of \mpm for the Korean language, \mpm-word. Instead of using the \emph{local POS prediction} module to predict a POS tag, the \mpm-word directly feeds the word embedding to an MLP, and use the output as the input of the \emph{masked POS reconstruction} module. Finally, we use the predicted tags after the Bi-LSTM layer as the induced tags.
The results are shown in Table~\ref{tab:context_app}. We do observe that the \mpm-word (width=2) variant, which is most similar to \citet{stratos2019mutual}, achieves the best result, demonstrating the effectiveness of such inductive biases on this Korean dataset.
Nonetheless, this preference is not consistent over languages. In our preliminary study, we notice many other languages still prefer our default model. We show the result on the 45-tag English dataset in Table~\ref{tab:context_app}, where the default \mpm show substantial advantages.
These preferences together with the observation in Appendix~\ref{appendix:mi} suggest different languages (or datasets) can prefer different types of dependency modeling (e.g. tag-tag dependency vs. word-tag dependency) and we encourage further study on this topic.

\section{Using Inducted Tags for Unsupervised Dependency Parsing}
\label{sec:app_parse}
In this section, as a side study, we test whether the performance trends on POS induction can transfer to unsupervised dependency parsing. We choose to use the Neural E-DMV model from \citet{jiang-etal-2016-unsupervised}, a commonly used baseline model that uses gold POS tags in the training. In our experiments, we replace the gold tags with the inducted tags from different models to see if the parsing performance correlates with the tag quality measured by M-1 accuracy. Following the convention in unsupervised parsing experiment setups, we train all the models on sections 2--21 of the English Penn WSJ dataset, and use section 22 for validation and section 23 for testing. We remove all the punctuation and only train and test on sentences with length $\leq$ 10 (i.e., following the WSJ10 setting). We compared three different models, our \mpm (78.6 M-1), the model from \citet{stratos2019mutual} (78.1 M-1), and the model from \citet{tran2016unsupervised} (75.0 M-1). 
We notice the models are highly sensitive to initialization. Hence, to remove the influence of bad initialization, we train each model ten different times and compare the best run. Using the gold tags, the E-DMV model can reach over 70 DDA (directed dependency accuracy). However, none of the models trained using predicted tags achieve DDA over 45, showing a substantial performance gap between the gold tags and the predicted tags. Surprisingly, while the tags from the Neural HMM model~\citet{tran2016unsupervised} have lower M-1 accuracy than the other two models, it shows a small advantage in the parsing performance over the over two models. We suspect the different trends may result from a mismatch between the objective optimized in parsing models and tagging models. The DMV objective explicitly models the transition probability between different nodes, hence the neural HMM model may have a slight advantage by using a more similar HMM-style objective.

\begin{table*}[t]
\centering
\small
\begin{tabular}{l ccc}
\toprule
& D(0) & MORPH & D(2-4)   \\
\midrule
\mpm (width=1) & \textbf{99.50} (95\%) & \textbf{95.41} (55\%) & 87.19 (0\%) \\
\ -connecting & 83.37 (0\%) & 85.09 (0\%) & 83.13 (0\%) \\
\mpm (width=2) & 92.99 (30\%) & 90.02 (30\%) & 87.62 (0\%) \\
\ -connecting & 83.97 (5\%) & 84.03 (0\%) & 83.44 (0\%)  \\
\mpm (full) & 96.50 (75\%) & 93.01 (30\%) & \textbf{95.31} (30\%) \\
\ -connecting & 75.02 (5\%) & 73.27 (0\%) & 82.81 (0\%)  \\
\midrule
\citet{stratos2019mutual} (width=2) & 92.99 (30\%) & 90.52 (30\%) & 86.56 (0\%) \\
\citet{tran2016unsupervised} (full) & 80.97 (0\%) & 82.54 (0\%) & 82.50 (0\%) \\
\bottomrule
\end{tabular}
\vspace{-5pt}
\caption{The Oracle M-1 score of different models on the synthetic dataset. The number in the bracket is the percentage of perfect runs (100 M-1).}
\vspace{-5pt}
\label{tab:syn_result_app}
\end{table*}

\section{Additional Agreement Learning Experiment Design and Results}
\label{sec:app_agreement}
\subsection{Additional Experiment Design}
Besides the D(0) and D(2-4) subsets introduced in Sec.~\ref{sec:agreement}, we add another variant: MORPH. 
The MORPH setting is a variant of D(0) with additional morphological features. While characters in every word in D(0) are all randomly generated, in MORPH, words with the n1 tags always end with -n1, words with the n2 tags always end with -n2, etc.
The tag-level regular expressions of all the subsets are shown in Table~\ref{tab:syn_design_app}.

\begin{figure}[t!]
    \centering
    \includegraphics[width=0.7\linewidth]{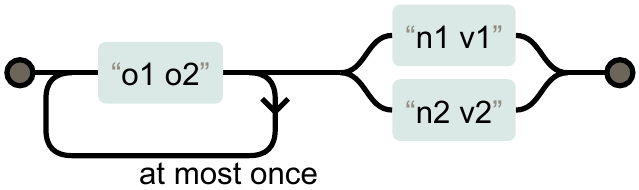}
    \caption{Illustration of the tag-level regular expression used to generate sentences for D(0) and MORPH. For MORPH, each word has useful morphological features, while all the characters in every word are randomly generated in D(0).}
\vspace{-3pt}
    \label{fig:syn_re_d0_app}
\vspace{-3pt}
\end{figure}

\subsection{Illustrations for Each Subset}
We provide illustrations of the tag-level regular expression for each subset. The illustration for D(2-4) is in Figure~\ref{fig:syn_re}. The tag-level regular expressions D(0) and MORPH are the same and the illustration can be seen in Figure~\ref{fig:syn_re_d0_app}.

\subsection{Additional Results}
We show additional results on the synthetic datasets in Table~\ref{tab:syn_result_app}.
Besides the results of the default \mpm, we also include an ablation of removing the ``Connecting $P(x|z)$ and $P(z|x)$'' trick introduced in Sec.~\ref{sec:additional_tech}. We can see connecting these two probabilities does bring substantial improvement on this agreement learning task.
Surprisingly, adding the morphological features (MORPH) does not help the models learn the agreement. Instead, nearly all models perform slightly worse on this variant. We suspect the problem may lies in the specific design of the morphological feature. The current setting provides an additional feature to first cluster n1 and n2 words, v1 and v2 words together since they have a common character `n' or `v' in the word, whereas normally we randomly sample characters to form the word. Hence, it can be easier for the models to enter the unideal local minimum.

\begin{figure}[t!]
    \centering
    \includegraphics[width=0.98\linewidth]{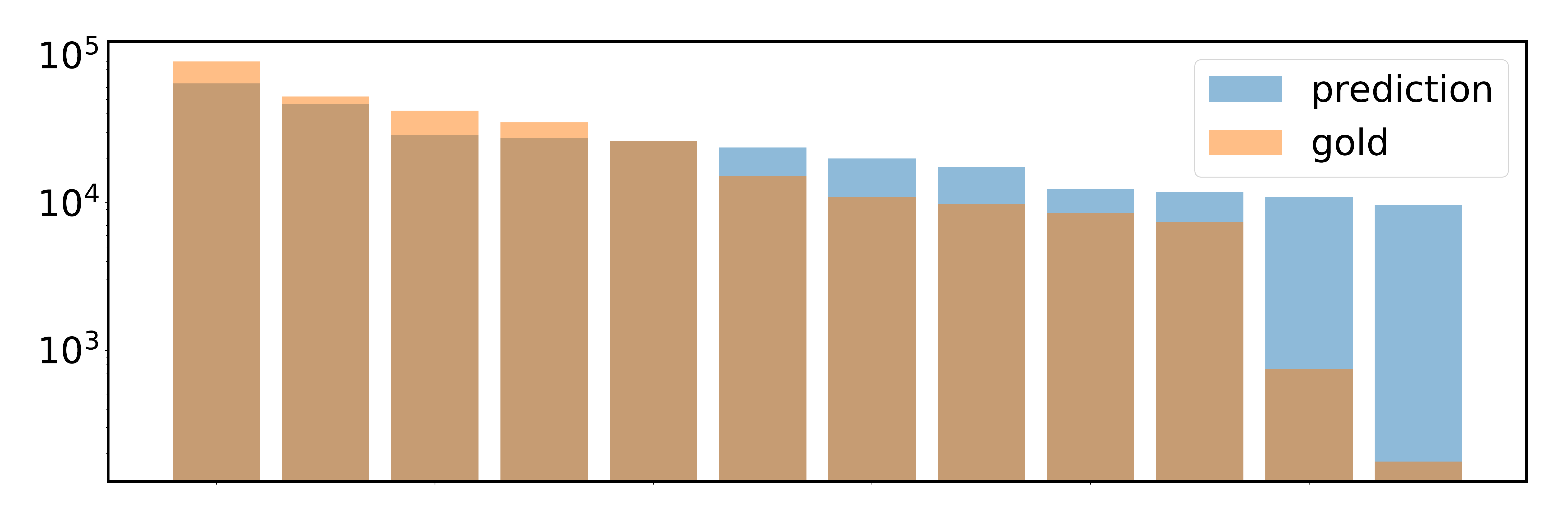}
    \caption{Log-scale sizes of the predicted clusters and the gold clusters for pt-br in the universal treebank.}
\vspace{-3pt}
    \label{fig:size_app}
\vspace{-3pt}
\end{figure}

\section{Predicted Clusters Analysis on Brazilian Portuguese}
\label{appendix:portuguese}

We provide additional analysis on the pt-br dataset in the universal treebank and check if our findings on the English dataset can generalize to the other language. Due to the 12-tag annotations on the universal treebank do not contain fine-grained tags, it is difficult to single out an agreement type to conduct a well-controlled analysis (like the subject-verb agreement analysis on English in the main paper), but below we verify all the other findings.

\paragraph{The sizes of predicted clusters are more uniform than gold clusters.}
Similar to the findings on the English 45-tag dataset, the sizes of predicted clusters are much more uniform than the gold clusters. A bar plot is shown in Figure~\ref{fig:size_app}.

\paragraph{Difficulty in mapping one word to multiple tags.}
Brazilian Portuguese also has words with different senses and POS-tags. For example, the word `parecido', which means `similar' in English, has three possible gold tags in the annotated data, including ADJ, VERB, and ADV. But again, in the model predictions, this word is always paired with the same tag.

\paragraph{Dataset biases influence predicted clusters.}
Since models on the universal treebank are only required to predict 12 tags, the influence of dataset biases is smaller than the English 45-tag data. However, we still can find some hints about the negative effect of the lack of linguistic diversity in the data. In the predicted clusters, nouns are separated into a number of clusters. Possibly due to the special domains of the data, the corpus includes lots of nouns representing locations and events. These nouns usually appear in a similar context after the ADP tag, hence models are likely to use a single cluster for these nouns, which is not ideal. While one can argue models should learn to separate the correct syntactic property from other spurious statistical properties, small datasets may not contain enough data to represent the complete picture of grammar. Hence, models are more likely to capture the statistical properties that are more common in the presented corpus and unlikely to induce the POS tags that ares more well-suited for the general language.

\section{Additional Experiment Results}
\label{appendix:others}
We briefly describe several variants we have tried in our preliminary experiments but \emph{do not} observe significant improvement. 
\begin{itemize}
    \item For the dependency modeling network in the masked POS reconstruction module in the \mpm model, we have also explored using a Transformer~\cite{vaswani2017attention} architecture or adding self-attention to the Bi-LSTM. However, we do not see substantial improvements.
    On the 45-tag English Penn Treebank dataset, our best Transformer result reaches 77.2 M-1, which is still lower than the average M-1 of the Bi-LSTM counterparts in Table~\ref{tab:main}.
    We suspect this trend is due to two reasons: (1) we notice that the Transformer models are more sensitive to initialization and hyper-parameter settings than the LSTM counterparts. Additionally, POS induction datasets are relatively small, which makes it harder to train a good Transformer model. (2) compared to Transformer models, LSTM models have the advantage of a preference of learning short-term dependencies first while learning long-term dependencies is still possible. This inductive bias could be useful for the POS induction task.

    \item Instead of directly training our model using gradient descent, another way to optimize our model is to view the reconstructed POS as latent variables and use EM-based algorithms to optimize the objective, similar to the method used in \citet{yang-etal-2019-latent}. However, in our experiments, we do not observe substantial improvement by using EM-based optimization methods over SGD-based methods.
\end{itemize}

\end{document}